\pdfoutput=1

\documentclass[11pt]{article}

\usepackage[preprint]{acl}
\usepackage{times}
\usepackage{latexsym}

\usepackage[T1]{fontenc}

\usepackage[utf8]{inputenc}

\usepackage{microtype}

\usepackage{inconsolata}

\usepackage{graphicx}

\usepackage{CJKutf8}
\usepackage{booktabs}
\usepackage{tabularx}
\usepackage{fancyvrb}
\usepackage{listings}
\usepackage{spverbatim}
\usepackage{subcaption}

\newcommand{\revise}[1]{\textcolor{black}{#1}}

\title{Automating Legal Interpretation with LLMs: \\Retrieval, Generation, and Evaluation}

\author{
    Kangcheng Luo$^{}$\thanks{These authors contributed equally to this work.},
    Quzhe Huang$^{}$\footnotemark[1],
    Cong Jiang$^{}$\footnotemark[1],
    Yansong Feng$^{}$\thanks{Corresponding author.} \\
    Wangxuan Institute of Computer Technology, Peking University\\
    Institute for Artificial Intelligence, Peking University and Peking University Law School\\
    {\tt luokangcheng@stu.pku.edu.cn} ~~
    {\tt huangquzhe977@gmail.com} \\
    {\tt \{jiangcong,fengyansong\}@pku.edu.cn}
}

\begin{document}
\begin{CJK*}{UTF8}{gbsn}

\maketitle
\begin{abstract}
Interpreting the law is always essential for the law to adapt to the ever-changing society. It is a critical and challenging task even for legal practitioners, as it requires meticulous and professional annotations and summarizations by legal experts, which are admittedly time-consuming and expensive to collect at scale. 
To alleviate the burden on legal experts, we propose a method for automated legal interpretation. Specifically, by emulating doctrinal legal research, we introduce a novel framework, \textbf{ATRIE}, to address Legal Concept Interpretation, a typical task in legal interpretation.
\textbf{ATRIE} utilizes large language models (LLMs) to \textbf{A}u\textbf{T}omatically \textbf{R}etrieve concept-related information, \textbf{I}nterpret legal concepts, and \textbf{E}valuate generated interpretations, eliminating dependence on legal experts. 
ATRIE comprises a legal concept interpreter and a legal concept interpretation evaluator. The interpreter uses LLMs to retrieve relevant information from previous cases and interpret legal concepts. 
The evaluator uses performance changes on Legal Concept Entailment, a downstream task we propose, as a proxy of interpretation quality.
Automated and multifaceted human evaluations indicate that the quality of our interpretations is comparable to those written by legal experts, with superior comprehensiveness and readability. Although there remains a slight gap in accuracy, it can already assist legal practitioners in improving the efficiency of legal interpretation.\footnote{The code and dataset are publicly available at \url{https://github.com/lkc233/ATRIE}}

\end{abstract}

\section{\revise{Introduction}}

Interpreting the law is always essential since laws are often vague \cite{endicott2000vagueness} and open-textured \cite{hart2012concept} to cover diverse real-world situations. For legal professionals, accurate interpretation is the foundation of fair judgments \cite{barak2005purposive}.  For laypeople, it determines whether they can understand and comply with the law, guiding their daily lives and decisions \cite{dworkin1982law}.  As shown in Figure~\ref{fig:process}, Theft in a dwelling is usually punished more severely than common theft. But what exactly is a “dwelling”? Is a school dormitory, tent, or motorhome a dwelling? Without clear interpretation, the law risks inconsistent application, undermining justice and public trust \cite{smits2017legal}. 

However, interpreting the law is far from easy. The main method that legal systems have developed is doctrinal legal research, which aims to provide clear, systematic, and well-reasoned legal interpretations \cite{tiller2006legal}.
Doctrinal legal research involves legal experts extensively reading a large volume of previous legal cases, books, papers, and other related materials to find valuable information \cite{Su2024}. Then, they summarize experience on detailed applications of the law. However, there are still
several challenges: (1) \textbf{Time-consuming:} Legal professionals must browse countless texts and cases to build a reliable interpretation. Despite advances in legal research tools, this remains a labor-intensive task that is not fully automated \cite{vangestel2011revitalizing}.  (2) \textbf{Untimely:} New cases continue to emerge at an increasing rate as society and technology progress. However, traditional methods rely on manual case-by-case reading to update interpretations, which is usually far behind judicial practice \cite{van2011methodologies}. (3) \textbf{Incomplete and Subjective:} Interpretations are limited by human capability. It is impossible to cover all existing cases, and interpretations remain incomplete.  Moreover, when selecting cases from the overall case pool, humans may unconsciously or even intentionally introduce their own biases~\cite{Farnsworth2011ImplicitBI}.

Previous studies have attempted to leverage LLMs for legal interpretation to mitigate the workload of human experts. \citet{hoffman2024generative} and \citet{engel2024asking} directly prompt LLMs for legal interpretation, asking LLMs to give the ordinary meaning of legal concepts. 
\citet{savelka2023explaining} utilize GPT-4 to interpret open-textured legal concepts from statutory articles based on expert-annotated valuable sentences from case law. These works have focused on interpreting \textbf{legal concepts}, as concepts represent the most important component of legal interpretation. Following prior research, we also select \textbf{legal concept interpretation} as the central focus of our work.  
However, previous works fail to address the above challenges because of the dependence on legal experts to (1) annotate concept-related valuable sentences from extensive volumes of case law and (2) evaluate the quality of LLM-generated legal concept interpretations.

Inspired by doctrinal legal research, we introduce \textbf{ATRIE} to \textbf{A}u\textbf{T}omatically \textbf{R}etrieve concept-related information, \textbf{I}nterpret legal concepts, and \textbf{E}valuate generated interpretations
without legal experts' intensive involvement. ATRIE comprises a legal concept interpreter and a legal concept interpretation evaluator.
The interpreter employs a Retrieval-Augmented Generation (RAG) framework \cite{lewis2020retrieval,guu2020retrieval}. It leverages LLMs to retrieve comprehensive and concept-related information from a vast database of previous cases, and then generates concept interpretations based on this information.
The evaluator is based on our proposed downstream task, called Legal Concept Entailment (LCE), which assesses models' understanding of legal concepts. Provided with different concept interpretations as references, the performance of a specific LLM on the LCE task serves as a proxy for the quality of concept interpretation. 
We recruit a legal expert to select 16 typical vague legal concepts and construct an LCE dataset to validate the effectiveness of our framework.
Our contributions are as follows:
\begin{itemize}
    \item We propose a novel automated framework for legal concept interpretation, which mimics doctrinal legal methods used by legal experts and eliminates experts' involvement.
    \item We introduce a downstream task, Legal Concept Entailment (LCE), together with a corresponding dataset, to automatically evaluate the quality of legal concept interpretations.
    \item  Automated and human evaluations
    demonstrate that our generated concept interpretations not only help LLMs better understand vague concepts but also 
    achieve high quality comparable to those written by legal experts.
    
\end{itemize}

\section{Related Works}

\subsection{Legal Interpretation}
Legal interpretation is the process of identifying the meaning of legal texts \cite{holmes1898theory}. It has been a longstanding challenge in the field of legal NLP \cite{nyarko2022statistical}. 
Initially, rule-based methods \cite{waterman1981models,paquin1991loge} provide users with tribunal decisions and doctrinal works to establish the meaning of open-textured legal concepts in specific contexts.
With the advancement of deep learning, researchs \cite{savelka2021discovering,savelka2021legal} use pre-trained language models to retrieve sentences from legal cases which are useful to explain legal concepts. 

With the rapid progress of LLMs, recent studies have also tried to use LLMs to interpret legal texts. \citet{jiang-etal-2024-leveraging} use LLMs to generate stories to make the law more accessible to the public. However, the story-based explanation is not precise enough to help legal professionals like lawyers or judges. \citet{coan2024artificial} use GPT to directly generate constitutional interpretation and \citet{engel2024professor} further add relevant cases to the input as references. These studies illustrate that using LLMs to interpret legal concepts is possible. However, they only evaluate one or two concepts. It remains uncertain whether their method could generalize to other concepts. \citet{savelka2023explaining} propose a general framework that could leverage valuable sentences from previous judgments to interpret legal concepts. 
It proves that augmenting the LLM with relevant sentences could improve the interpretation quality 
and eliminate the issue of hallucination. 
However, its valuable sentences are manually selected from judgments, which is costly.

Previous approaches depend on legal experts to annotate concept-related information and evaluate generated interpretations. This reliance on manual expert input prevents them from addressing the aforementioned challenges. To overcome this, we introduce an automated framework that retrieves concept-related information, interprets legal concepts, and evaluates the resulting interpretations.

\subsection{{Doctrinal Legal Research}}

Doctrinal legal research, which dates back to the Roman Empire~\cite{van1998legal}, is a traditional and systematic method of analyzing legal texts, case law, and principles~\cite{pound1913end}. The goal is to identify, explain, and systematize legal rules, and sometimes to predict how courts might apply them in the future~\cite{tiller2006legal}. This method is widely accepted in courts and legal scholarship as the standard way to work with legal texts around the world~\cite{bhat2019idea}. Many legal tasks, including legal interpretation, mainly depend on doctrinal legal research~\cite{majeed2023doctrinal}. Doctrinal legal research provides a clear approach to analyze the meaning of unclear or vague legal texts. Traditionally, this work is done by human experts. It often relies on significant human effort to review extensive cases, identify patterns, and extract insights. This process is time-consuming and labor-intensive~\cite{hutchinson2012defining}. \citet{Su2024} suggests that the task within doctrinal legal research of reading cases and extracting theoretical insights could potentially be performed by Legal AI. Inspired by the process of doctrinal legal research, our work explores how LLMs can automate legal interpretation, making it faster and more scalable.

\section{Task Definition}

\revise{In this work, we rely exclusively on previous cases as reference materials to interpret vague concepts in the articles. We use cases because they are the most concrete and fundamental sources. Books and papers often cite cases to support their arguments.}

We formally define the task as follows: Given a legal article $a$ and a vague concept $c$ contained within, the objective is to generate a legal interpretation $e$ for $c$ based on previous cases, clarifying the conditions of its applicability.
For example, as shown in Figure~\ref{fig:process}, our goal is to address the vague concept of "Dwelling" within Article 264 of the Criminal Law. We aim to leverage previous cases to interpret the circumstances under which "Dwelling" applies.

\section{Legal Concept Interpreter}
\label{method}

\begin{figure*}[!t]
	\centering
	\includegraphics[width=1\linewidth]{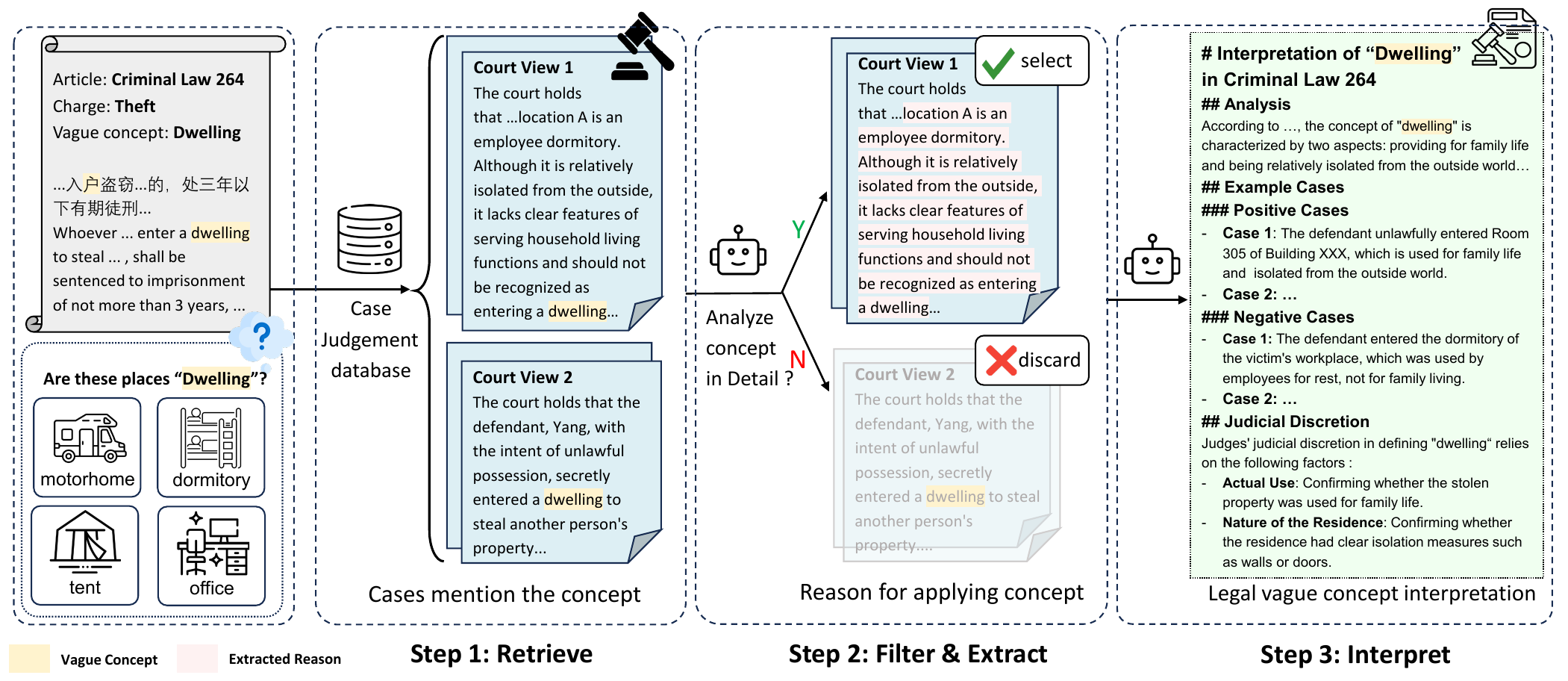}
 \caption{Overview of our legal concept interpreter.}
  \vspace{-2ex}
 \label{fig:process}
\end{figure*}

Following the method of legal experts, our legal concept interpreter summarizes the detailed applications of a given vague concept in judicial practice based on relevant cases. Specifically, it is composed of three parts (Figure~\ref{fig:process}):
(1) \textbf{Retrieve}: Retrieve cases that mention the concept.
(2) \textbf{Filter\&Extract}: Select cases where the concept is analyzed in detail within the cases and extract the reasons why the concept applies or not.
(3) \textbf{Interpret}: Use LLMs to generate the interpretation of the concept based on the extracted reasons.

\subsection{Retrieving Cases}
\label{retrieve}

To find cases helpful to interpret the vague concept, the first step is to retrieve those \textit{mention} the concept. Formally, given a vague concept $c$ and the article $a$ to which $c$ belongs, we find all the cases citing the number of article $a$ from a case database. Then, we retrieve the cases that mention concept $c$ through exact string matching. All the retrieved cases form case set $\mathcal{D}_0$.

Our case database is constructed by collecting \textit{judgment} documents published on China Judgments Online\footnote{\href{https://wenshu.court.gov.cn/}{https://wenshu.court.gov.cn/}}.  
It's the largest public case platform in China and the official website hosted by the Supreme People’s Court of China. Our database includes cases from 1985 to 2021, which ensures the source's comprehensiveness.

To further refine this retrieval process, we focus specifically on the most relevant section of the case document. A case typically contains five parts: Header, Facts, Court View, Verdict, and Conclusion \footnote{Details of the case structure are in Appendix \ref{judgment_introduction}.}. Among them, the court view section explains the legal rationale and basis for the judgment. 
We use exact string matching to retrieve the cases that contain the vague concept in their court views. Legal terminology demands precision with fixed expressions that rarely permit alternative phrasings, so this approach ensures accuracy over fuzzy matching methods like dense retrieval.

\subsection{Filtering Relevant Cases and Extracting Reasons}
\label{filter}
In this step, we filter \textit{relevant cases}—defined as those in which the court view sections provide detailed reasons why the vague concept applies to the case or not—and extract the reasons.
This filtering is necessary because cases lacking such detailed discussion offer few valuable insights for generating interpretations.
\footnote{We show an example of a case that mentions the concept only and a relevant case that discusses the concept in detail in Appendix~\ref{appendix:relevant_case}.}

First, we use LLMs to filter the relevant cases from $\mathcal{D}_0$.\footnote{All the prompts we use are shown in Appendix~\ref{appendix:prompt}.}
Taking the court view as input, we require the LLM to determine whether it provides a detailed reason $r$ and extract this reason if provided.
The reason $r$ should be a combination of original sentences from the court view. Next, we prompt LLMs to determine whether the concept applies to the case based on the court view, yielding a binary label $l$ (Yes/No). 
From this process, we obtain a refined case set $\mathcal{D}_1$ containing cases that discuss the concept in detail in the court view.

Upon analyzing the labels within $\mathcal{D}_1$, we observe the proportion of positive cases (where $c$ applies to the case) far exceeds negative cases, with a ratio surpassing 10:1. 
This phenomenon could potentially be attributed to the exclusive inclusion of prosecuted and adjudicated cases in China Judgments Online. In judicial practice, only cases with substantial evidence supporting the prosecution are brought to court. As a result, the concept is more likely to apply to these cases, which leads to a higher proportion of positive examples. To comprehensively account for different situations when generating concept interpretations, we aim to ensure that both positive and negative examples receive adequate attention. Therefore, we only sample a subset of positive cases to construct a balanced dataset $\mathcal{D}$ and its corresponding reason set $\mathcal{R}$.

\subsection{Generating Concept Interpretations}
\label{sec:generate}

After collecting relevant cases and reasons, this step leverages an LLM to summarize these past experiences and generate an interpretation of the vague concept. 

An interpretation should elaborate on how courts have explained or applied the vague concept. We design the interpretation to consist of three main components (see Appendix \ref{appendix:interpretation_structure}): \textit{Analysis}, which explains the basic meaning of the concept and its applicability conditions; \textit{Case Examples}, which provides representative positive and negative cases from past rulings; and \textit{Judicial Discretion}, which offers criteria to guide judges in flexibly applying vague concepts based on case specifics. 

The input to the LLM for generating interpretations consists of the following components: (1) legal article $a$, (2) vague concept $c$, (3) reason set $\mathcal{R}$, and (4) interpretation example $e_0$. We require the output interpretation to follow the same format as the interpretation example $e_0$ to ensure a consistent and standardized format (Appendix \ref{appendix:e_0}).

\section{Legal Concept Interpretation Evaluator }

Previous work~\cite{savelka2023explaining} has predominantly relied on human evaluation to evaluate the quality of the generated interpretations. We also conducted human evaluations, as detailed in Section \ref{sec:human_score}. However, human evaluation is inherently subjective, and we aim to assess the quality of the generated concepts more objectively and quantitatively. Therefore, we design the legal concept interpretation evaluator based on a new task we propose, Legal Concept Entailment. It enables an objective and reproducible comparison of different interpretations' quality.

\subsection{Legal Concept Entailment}

\begin{figure}[!t]
	\centering
	\includegraphics[width=1\linewidth]{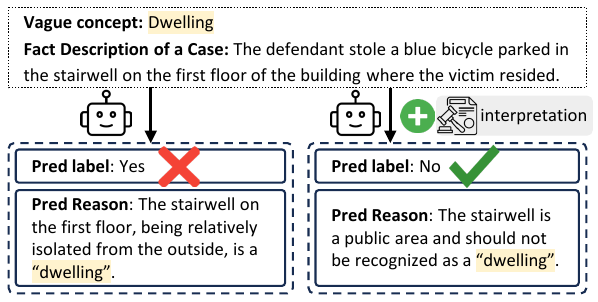}
 \caption{An example of Legal Concept Entailment Task. The left half of the figure illustrates the LLM directly performing the task, while the right half shows the LLM completing the task with the concept interpretation as a reference.}
  \vspace{-2ex}
 \label{fig:entailment_example}
\end{figure}

If an interpretation of a concept is effective, it should help humans or models better determine whether the concept applies to unseen cases. Based on this assumption, we design the downstream task Legal Concept Entailment (LCE). Given the fact description of a case relevant to the vague concept, the task is to determine whether the concept applies and provide a reason. We use a fixed LLM
to perform this classification task. By incorporating different interpretations into the input, we can observe changes in the classification accuracy, which allows us to assess the quality of the interpretations. More accurate classification demonstrates higher-quality interpretation.

The LCE task is divided into two parts. The first part is a binary classification task. For a vague concept $c$ in a legal article $a$, given the fact description $f$ of an unseen relevant case $d$, the output should be a binary label $\hat{l}$ (Yes/No), indicating whether $c$ applies to the fact $f$. The second part is a generation task, which requires generating a reason $\hat{r}$ to explain the prediction result of the binary classification task. An example is shown in Fig \ref{fig:entailment_example}.

\subsection{LCE Dataset}
\label{dataset}

\revise{We recruit a legal expert with extensive judicial experience to identify 16 vague legal concepts in 14 legal articles (Appendix~\ref{appendix:concepts}). These concepts are representative and frequently used in judicial practice.
The statistical analysis reveals that, among all the cases in our database that cite these legal articles, 24.9\% involve the corresponding vague concepts.
Thus, we leverage them to demonstrate the effectiveness of our framework.}

For each concept, we reuse the retrieval and filtering modules described in Sec~\ref{retrieve} and~\ref{filter} to collect relevant cases. These cases are considered challenging because the court view provides detailed explanations of the vague concepts. On average, 166 cases are selected for each concept, with a positive-to-negative case ratio 2:1. 
Detailed statistics are provided in Appendix~\ref{appendix:concepts}.

Following methods outlined in Sec \ref{filter}, we use Qwen2.5-72B-Instruct~\cite{qwen2.5} to annotate each case with the gold label $l$ and reason $r$ for LCE task. Manual inspection indicates that the annotated data is highly accurate (Appendix \ref{appendix:manual_inspection}).

The distinction between data annotation and LCE task lies in the input provided to the LLM. For annotation, the input is the court view, which contains explicit judgments made by judges and can be directly extracted as ground truth. In contrast, for the LCE task itself, the input is the fact description, which lacks explicit judgments, requiring the LLM to perform reasoning to infer the entailment.

\subsection{Evaluation Metrics}
For the classification task, we use Accuracy (Acc.), Macro Precision (Ma-P), Macro Recall (Ma-R), and Macro F1 (Ma-F) as the evaluation metrics. The use of the macro average is motivated by the imbalance in the number of cases relevant to each concept, to assign equal weight to all concepts.

For the reason generation task, we use an LLM-based evaluator to evaluate the consistency between the generated reason $\hat{r}$ and the gold reason $r$ from the court view, following previous LLM-as-a-Judge based methods~\cite{zheng2023judging,zhu2023judgelm}. 
\revise{In our main experiments, we use GPT-4o~\cite{achiam2023gpt} as the evaluator. However, we find that open-source LLMs, such as Qwen2.5-72B, produce highly consistent evaluation results with GPT-4o (Appendix \ref{appendix:llm_cs}), suggesting they can serve as a viable substitute.}
We require GPT-4o to rate from 1 to 10 for the consistency between the $\hat{r}$ and $r$, with higher scores indicating greater consistency. 
Note that the consistency score is directly set to 0 if the classification result is incorrect.

\subsection{Evaluation Process}

This section describes how our evaluator assesses the interpretations generated by our framework.

First, we generate the interpretations to be evaluated using our legal concept interpreter. To prevent data leakage, the cases used for generating interpretations do not overlap with the test dataset. Next, we prompt the LLM to perform the LCE task using the generated interpretations.

As shown in the right half of Figure \ref{fig:entailment_example}, given a vague concept $c$ in a legal article $a$ and the fact description $f$ of a relevant case $d$, the LLM is prompted to analyze whether the concept $c$ applies to the fact $f$ based on the concept interpretation. Specifically, the LLM first generates a reason $\hat{r}$ and subsequently assigns a classification label $\hat{l}$.

\subsection{Baselines}
\label{sec:baseline}

We compare our method with two baseline categories: "w/o Interpretation," in which the LLM relies solely on its internal knowledge, and "w/ Interpretation," in which the LLM is provided with an interpretation of the vague concept for the task.

\paragraph{w/o Interpretation}
(1) \textbf{Random}: We use random guessing of "Yes" or "No" as a weak baseline.
(2) \textbf{Zero-shot (ZS)}: The LLM performs the LCE task in a zero-shot setting. Specifically, only the legal article $a$, the vague concept $c$, and the fact description $f$ of the relevant case $d$ are provided as input. (Shown in the left half of Figure \ref{fig:entailment_example}.)
(3) \textbf{Chain-of-Thought} \cite{kojima2022}: 
Using the prompt "Let’s think step by step" to encourage the LLM to generate intermediate steps and improve its reasoning.

\paragraph{w/ Interpretation}

We introduced concept interpretations generated by different approaches, including human-written and LLM-generated interpretations:
(1) \textbf{Judicial Interpretation (JI)}: We recruit a legal expert to retrieve judicial interpretations
for the concept \textit{c}.
Judicial interpretations are explanations issued by the Supreme People's Court on how to apply the law specifically.
(2) \textbf{Expert interpretation (EI)}: We collect legal professionals' interpretations for the concept \textit{c} from FaXin\footnote{
\href{https://www.faxin.cn/}{https://www.faxin.cn/},}  
and WeChat official accounts of major law firms, which are of high quality. 
(3) \textbf{LLM Direct Interpretation (DI)}
: Without providing relevant cases, the LLM generates an interpretation of the vague concept $c$ directly based on its internal knowledge.

\subsection{Implementation details}
\label{appendix:implement}

After filtering, we obtained 2,642 cases and extracted the same number of reasons for generating concept interpretations. On average, each concept was associated with 165 cases.
We use the open-source LLM Qwen2.5-72B-Instruct with a maximum context length of 128k tokens to generate vague concept interpretations. The temperature is set to 0.9 to encourage more diverse outputs. Detailed prompt information can be found in Appendix~\ref{appendix:prompt_generate}. 

To investigate the effectiveness of our generated interpretations in assisting models with different capabilities, we employ Qwen2.5-72B-Instruct and Qwen2.5-14B-Instruct to perform the LCE task. 
We use gpt-4o-2024-08-06\cite{achiam2023gpt} to give the consistency score.
To reduce the randomness of the output, the temperature of all LLMs in LCE task is set to 0.

\subsection{Result}

\begin{table*}[h]
  \centering
   \scalebox{0.78}{
  \begin{tabular}{c|ccccc|ccccc} 
     \hline
    & \multicolumn{5}{c}{Qwen2.5 (72B)}
& \multicolumn{5}{c}{Qwen2.5 (14B)} 
\\ 
 \cline{2-11}
 & Acc & Ma-P & Ma-R & Ma-F & CS
& Acc & Ma-P & Ma-R & Ma-F & CS
\\ 
 \hline
    Random
    & 51.66 & 51.13 & 51.23 & 50.32 & /
    & 51.66 & 51.13 & 51.23 & 50.32 & / \\ 
    Zero-Shot
    & 71.38 & \underline{72.64} & 61.81 & 61.42 & 5.658
    & 70.92 & \underline{73.04} & 60.78 & 59.88 & 5.525\\ 
    Chain-of-Thought
    & 71.95 & 72.07 & 63.26 & 63.46 & \underline{5.717}
    & 71.52 & \textbf{73.83} & 61.60 & 61.01 & 5.666\\ 
    \hline
    
    Judicial Interpretation
    & 72.10 & 69.87 & 65.82 & 66.54 & 5.573
    & 70.92 & 68.24 & 64.62 & 65.23 & 5.347\\ 
    Expert Interpretation
    & 72.13 & 70.78 & 64.68 & 65.30 & 5.630
    & 71.95 & 69.85 & 65.31 & 66.01 & 5.581\\ 
    Direct Interpretation
    & \underline{72.35} & 70.03 & \underline{66.43} & \underline{67.18} & 5.642
    & \underline{72.72} & 70.98 & \underline{66.11} & \underline{66.90} & \underline{5.677}\\ 
    \hline 
    ATRIE
    & \textbf{75.03} & \textbf{73.21} & \textbf{69.97} & \textbf{70.87} & \textbf{5.946}
    & \textbf{74.50} & 72.49 & \textbf{69.56} & \textbf{70.39} & \textbf{5.840}\\

 \hline
  \end{tabular}
  }
  \caption{\label{tab:Main-result}
    Main results of automated evaluation on the Legal Concept Entailment task, the best is \textbf{bolded} and the second is \underline{underlinded}. CS represents the consistency score. We use Qwen2.5-72B to generate concept interpretations and employ Qwen2.5-72B/14B to perform the LCE task.
  }
\end{table*}

We report the performance of our method and all baselines on the LCE Task in Table~\ref{tab:Main-result}. 
Overall, ATRIE achieves the best performance across nearly all models, showcasing the effectiveness of our framework and the necessity of its core components.

\subsubsection{Classification Task}
For the classification task, we found that: 
\paragraph{(1) LLMs exhibit a certain degree of discriminative ability by leveraging their internal knowledge.}
The performance of the "w/o Interpretation" setting surpasses that of random guessing, yet there remains significant room for improvement. This suggests that using the LLM's performance on the LCE task as a proxy for interpretation quality is a feasible approach.
\paragraph{(2) Interpretations for vague concepts are valuable.}
 "w/ Direct Interpretation" outperforms all the "w/o Interpretation" settings, showing that LLMs can leverage their extensive internal knowledge to reason about vague concepts and generate useful legal concept interpretations. "w/ Judicial Interpretation" falls short of "w/ Direct Interpretation." We attribute this to the relatively simple explanations provided in judicial interpretations, which lack the depth required to guide LLMs in evaluating the applicability of vague concepts to specific cases. The performance of "w/ Expert Interpretation" is inferior to ATRIE. We attribute this to the fact that expert-written interpretations are often overly abstract and detailed, which results in poorer readability. We will further discuss this in the human evaluation (Sec~\ref{sec:human_eval}).
\paragraph{(3) Utilizing relevant cases is necessary for good concept interpretations.}
ATRIE outperforms "w/ Direct Interpretation", demonstrating the effectiveness of referencing relevant cases in generating interpretations.

\subsubsection{Reason Generation Task}

For the reason generation task, we found that: 
(1) the consistency score of ATRIE is the highest, showing a significant improvement over both "w/o Interpretation" and "w/ Interpretation" baselines. This indicates that the interpretations generated by our method help the model better understand the concepts and make correct inferences.
(2) Other "w/ Interpretation" methods generally perform worse than CoT despite showing improvements in classification tasks. We contend that this arises from these interpretations being incomplete or including irrelevant information, which misguides the LLM to reason in an incorrect direction.

\subsection{Case Study}
\begin{figure*}[h]
	\centering
	\includegraphics[width=0.9\linewidth]{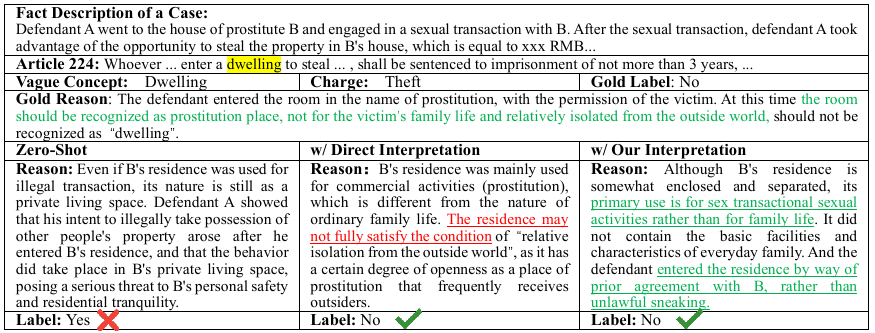}
 \caption{Case study for Legal Concept Entailment Task. The \textcolor{green}{\underline{green}} parts indicate reasoning processes consistent with the gold reason, while the \textcolor{red}{\underline{red}} parts indicate inconsistencies.}
  \vspace{-2ex}
 \label{fig:case_study_1}
\end{figure*}

Figure \ref{fig:case_study_1} presents an example of different methods applied to the LCE Task. As demonstrated in the case, our interpretation accurately understands the applicability conditions of "dwelling" and outputs the correct prediction with the right reasoning path. 
In contrast, Zero-shot gave an incorrect answer due to a misunderstanding of the concept of 'dwelling'. For "w/ Direct Interpretation", although it reaches the correct conclusion, the reasoning process contains errors and uncertainties. In this scenario, it failed to clarify the vague concept, using the expression "may not fully satisfy".

\section{What Affects Interpretation Quality?}

This section discusses the impact of different settings in our interpreter on the quality of generated interpretations.

\subsection{Retrieval Method}

\begin{table}[h]
  \centering
   \scalebox{0.75}{

  \begin{tabular}{c|cc}
    \hline
    &  \multicolumn{2}{c}{Qwen2.5 (14B)} \\
  & Ma-F & CS   \\
    \hline
    No Retrieval & 66.90 & 5.677 \\
    String Match  & 69.04 & 5.772 \\
    String Match + Filter & 69.60 & 5.817 \\
    String Match + Filter + Balance (ATRIE) & \textbf{70.39} &  \textbf{5.840} \\
    \hline
  \end{tabular}
}
  \caption{Ablation study for relevant case retrieval.}
  \label{tab:retrieval_methods}
\end{table}

To verify the importance of each step in our process of retrieving relevant cases, we compare the performance of the following settings: 
(1) \textbf{No retrieval} does not retrieve cases (i.e., LLM Direct Interpretation); (2) \textbf{String Match} does not use LLM to filter cases or perform label balancing; 
(3) \textbf{String Match + Filter} only removes the label balancing stage.
We ensure that the number of cases retrieved by each method is consistent. 
Table~\ref{tab:retrieval_methods} shows that every component of our retrieval method is necessary.

\subsection{Number of Cases}

\begin{figure}[!t]
	\centering
	\includegraphics[width=0.85\linewidth]{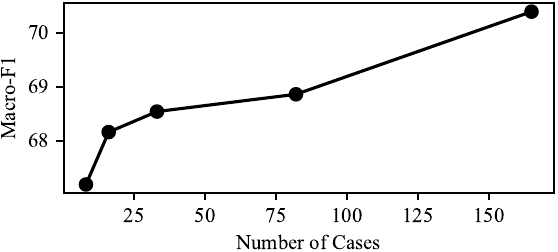}
 \caption{Results of different numbers of cases utilized to generate the interpretations. The model for generating interpretations and the prediction model are Qwen2.5-72B and Qwen2.5-14B, respectively.}
  \vspace{-2ex}
 \label{fig:case_num}
\end{figure}

We investigated the impact of using different numbers of cases on the quality of generated concept interpretations. Specifically, we sampled different numbers of reasons from the extracted reason set $\mathcal{R}$ as input to the LLM. Figure  \ref{fig:case_num} shows that more input reasons lead to higher-quality interpretations.

The more cases legal experts review, the more comprehensive their concept interpretations become. Our findings align with legal experts' experiences, showcasing LLMs' ability to analyze numerous cases effectively and highlighting their advantage in aiding legal interpretation.

\subsection{Which parts of a case are useful?}
In Section~\ref{filter}, we only extract a few sentences discussing the concept from the court view of each relevant case without including the complete fact description and court view. We aim to investigate whether this might result in the loss of important information from the case, potentially affecting the generation of interpretations. To explore this, we compared three different approaches to representing the relevant information in a case during the interpretation generation step:
(1) \textbf{Court View}: the part of the judgment where the judge explains the legal rationale and interprets the basis of the ruling;\ (2) \textbf{Summarized Fact and Court View}: 
The fact section in a case is often lengthy and contains excessive detail. To address this, we first use Qwen2.5-72B-Instruct to summarize the facts and then concatenate it with the court view section; (3) \textbf{Extracted Reason}: Extracted reasons in Section~\ref{filter}.

\begin{table}[h]
\centering
    \scalebox{0.8}{
      \begin{tabular}{p{4.2cm}|>{\centering\arraybackslash}p{1.5cm}>{\centering\arraybackslash}p{1.5cm}}
        \hline
        & \multicolumn{2}{c}{Qwen2.5 (14B)} \\
         & Ma-F & CS \\
        \hline
        Court View  & 69.10 & 5.775 \\
        Fact \& Court View  & 70.17 & 5.818 \\
        Extracted Reason (ATRIE) & \textbf{70.39} & \textbf{5.840} \\
        \hline
      \end{tabular}
    }
    \caption{Results of using different parts of cases to generate interpretations.}
    \label{tab:retrieval_parts}
\end{table}

In the experiment, we control the number of input cases to be the same. In practice, using the "Extracted Reason" allows for the inclusion of more cases, as each entry is shorter in length. Even in this scenario with the same number of cases, Table \ref{tab:retrieval_parts} shows that "Extracted Reason" performs best, indicating that it retains vital information while filtering out redundant details.

\subsection{Components of Interpretation}
In Section \ref{sec:generate}, we ask the model to output the following components: Analysis, Example Cases, and Judicial Discretion. We aim to investigate whether each component is necessary.
Specifically, we delete one main component at a time while keeping the other parts unchanged.

The results (Table \ref{tab:ablation}) show that each component of the generated concept interpretation contributes to the overall performance. Removing the "Example Cases" section results in the most significant performance drop, highlighting the importance of providing specific case examples.

\begin{table}[h]
  \centering
   \scalebox{0.8}{ 
  \begin{tabular}{l|c}
    \hline
    & \multicolumn{1}{c}{Qwen2.5 (14B)} \\ 
    &  Macro-F1  \\
    \hline
    w/o Example Cases & 67.41  \\
    
    \quad - w/o Positive Cases           & 68.17  \\
    \quad - w/o Negative Cases           & 69.98  \\
    
    w/o Analysis                 & 70.43  \\
    w/o Judicial Discretion      & 70.69  \\
    \hline
     ATRIE   & \textbf{70.87} \\
    \hline
  \end{tabular}
}
  \caption{Results of ablation experiments on different components of generated concept interpretations.}
  \label{tab:ablation}
\end{table}

\subsection{\revise{Are Legal LLMs More Effective?}}
Previously, we utilized general-purpose LLMs for generating legal interpretations. However, we also aimed to evaluate the performance of legal LLMs.
ATRIE requires analyzing hundreds of cases, with an average input length of 17k tokens. In contrast, among the currently available Chinese legal LLMs, Farui-plus\footnote{\href{https://tongyi.aliyun.com/farui}{https://tongyi.aliyun.com/farui}}—which offers the longest maximum context length—supports only up to 12k tokens (Appendix~\ref{app:legalllm}). Thus, we restrict the input length to within 10k tokens and compare the concept interpretations generated by Farui-plus and Qwen2.5-72B-Instruct under identical input conditions. Table~\ref{tab:legal_llm} shows that general-purpose LLM Qwen significantly outperforms legal LLM Farui in interpreting legal concepts. This phenomenon may be attributable to the superior long-text comprehension and generation capabilities of Qwen2.5-72B-Instruct, which enables more effective summarization from extensive texts.

\begin{table}[h]
    \centering
    \scalebox{0.75}{
    \begin{tabular}{c|ccccc}
        \hline
        & \multicolumn{5}{c}{Qwen2.5 (14B)} \\
         & Acc & Ma-P & Ma-R & Ma-F & CS \\
        \hline
        Zero-Shot & 70.92 & \textbf{73.04} & 60.78 & 59.88 & 5.525 \\
        ATRIE (Farui) & 72.02 & 70.35 & 64.86 & 65.51 & 5.630 \\
        ATRIE (Qwen) & \textbf{73.27} & 72.86 & \textbf{67.60} & \textbf{68.45} & \textbf{5.736} \\
        \hline
    \end{tabular}
    }
    \caption{Evaluation results of concept interpretation generated by Farui-plus and Qwen2.5-72B.}
    \label{tab:legal_llm}
\end{table}

\section{Human Evaluation}
\label{sec:human_eval}

In this section, we further analyze the strengths of our interpretations through human evaluation.
\label{sec:human_score}
\subsection{Evaluation Metrics}

\revise{We recruited 2 legal experts who have passed China's Unified Legal Profession Examination to assess the legal concept interpretations generated by Qwen2.5-72B-Instruct. They collaboratively establish five evaluation criteria and score the interpretations}: 
(1) \textbf{Accuracy (Acc.)}, 
(2) \textbf{Informativeness (Info.)},
(3) \textbf{Normativity (Norm.)}, 
(4) \textbf{Comprehensiveness (Comp.)}, 
(5) \textbf{Readability (Read.)}. 
We use a 10-point Likert scale, where 1 represents "very poor" and 10 represents "very good". 
\footnote{Details 
about the metrics and human evaluation 
are
discussed in Appendix \ref{appendix:human_eval}.}

\subsection{Results}
\revise{We compare three different interpretations in Sec~\ref{sec:baseline} for each of the 16 legal concepts.} In the Appendix~\ref{appendix:human_eval_case_study}, we present a case study to illustrate a comparative analysis of the scores for interpretations generated by two distinct methods, along with the rationale behind these scores.
In Table \ref{tab:human_eval}, we have several
observations: 
(1) The average score of ATRIE is the highest, indicating that our interpreter can generate legal concept interpretations comparable to those produced by legal experts.
(2) The Comprehensiveness score of ATRIE is much higher than Expert Interpretation, indicating that having LLMs read a vast number of cases helps generate more comprehensive concept interpretations.
(3) Expert Interpretation (EI) receives the lowest score in Readability, indicating that the interpretations written by legal experts tend to be abstract or complex, which may hinder understanding by both humans and LLMs.
(4) In Accuracy, Informativeness, and Normativity, ATRIE shows improvements over Direct Interpretation (DI). Although there are still minor gaps between ATRIE and Expert Interpretation, it's important to note that Expert Interpretation was produced by legal experts who spent considerable time. 

\revise{In addition, experiments on efficiency (Appendix \ref{app:efficency}) demonstrate that ATRIE significantly reduces both time and money costs for concept interpretation generation compared to legal experts.}
In the future, combining the two approaches may be a better option. Legal experts can revise a draft generated by the LLM to improve efficiency.

\begin{table}[h]
  \centering
   \scalebox{0.75}{

  \begin{tabular}{ccccccc}
    \toprule
    & Acc. & Info. & Norm. & Comp. & Read. & Avg.\\
    \midrule
    DI &  7.03 & 6.21 & 7.53 & \underline{6.72} & \textbf{7.38} & 6.97  \\ 
    EI &  \textbf{7.68} & \textbf{7.03} & \textbf{8.00} & 6.12 & 6.26 & \underline{7.02}  \\ 
    ATRIE & \underline{7.18} & \underline{6.76} & \underline{7.76} & \textbf{7.15} & \underline{7.18} & \textbf{7.21}  \\
    \bottomrule
  \end{tabular}
}
  \caption{
    Human evaluation results of vague concept interpretations. 
    "Avg." represents the average score across five evaluation metrics.
  }
  \label{tab:human_eval}
\end{table}

\section{Conclusion}
In this work, we explore the use of LLMs to address a challenging task in the legal field: Legal Interpretation. By emulating doctrinal legal research, we propose a fully automated framework for retrieving concept-related information, interpreting legal concepts, and evaluating the generated interpretations.
Both automated and human evaluations demonstrate that our generated interpretations are useful and comparable to those written by legal experts. Our study suggests considerable potential for using LLMs to assist legal experts in legal interpretation and beyond.

\section*{Limitations}
\paragraph{\revise{Sample Size}}
We merely use 16 typical vague concepts as examples to demonstrate our framework's effectiveness and build a usable dataset for the LCE task. Actually, our method can explain any concept as long as it has been applied in legal practice and is supported by a sufficient number of cases. 
However, in China and other countries such as Switzerland, the judicial system only discloses a very small portion of cases. Within these limited publicly available cases, the selected 16 concepts by legal experts are typical; thus, there is a sufficient number of released relevant cases. As judicial systems internally possess the entire database of cases, our method holds significant potential for application within the court or other institutions, offering substantial assistance to judges and other legal practitioners.
\paragraph{Potential Risk of Data Leakage}
Although the LLMs used in our experiments on the LCE task are open-source, their training dataset is not fully transparent, which raises the possibility of data leakage. To address this issue, we evaluated different interpretations using the same LLM to ensure a fair comparison. The relative performance changes on the LCE task demonstrate our advantages.
\section*{Ethical Considerations}

\paragraph{Privacy and Data Security}
Legal datasets frequently contain sensitive details about individuals and organizations, and improper handling can result in significant privacy violations. To safeguard this information, the case dataset used in our experiments is thoroughly anonymized.

\paragraph{\revise{LLM-Related Risks}}
Large language models (LLMs) can inherit biases or inaccuracies from the data they are trained on, potentially leading to flawed legal interpretations. While LLMs can assist in generating legal concepts, they should not replace human judges or be used directly in real-world decision-making. Human oversight is essential to ensure fairness and accuracy in legal processes. 

Despite this, we would like to clarify that our framework does not pose serious risks when applied to real cases; instead, it provides significant assistance to judges.

First, our method focuses on interpreting legal concepts rather than delivering final judgments. The ultimate decision-making authority remains with the judge. In real-world applications of LLM technology in law, these models serve only as auxiliary tools, while accountability still rests with human judges~\cite{liu2024judges}.

Second, even legal experts may have differing or sometimes incorrect interpretations. Whether reading AI-generated explanations or those written by legal professionals, judges and lawyers always verify the information themselves. Therefore, AI does not introduce greater risks but instead significantly reduces the time required to review cases. Legal professionals have the expertise to assess and identify potential flaws in interpretations.

\paragraph{Code of Conduct}
This research follows the ACL Code of Ethics and respects participants’ anonymity. We obtain the consent of two legal experts who passed China's Unified Qualification Exam for Legal Professionals and recruit them for manual annotation and experiments. We pay them wages higher than the local average hourly rate and ensure that the content generated by the LLM is safe and non-offensive.

\section*{Acknowledgements}
This work is supported in part by NSFC (62161160339) and Beijing Science and Technology Program (Z231100007423011). We thank the anonymous reviewers for their valuable suggestions. For any correspondence, please contact Yansong Feng.
\bibliography{custom}

\begin{thebibliography}{40}
\providecommand{\natexlab}[1]{#1}

\bibitem[{Achiam et~al.(2023)Achiam, Adler, Agarwal, Ahmad, Akkaya, Aleman, Almeida, Altenschmidt, Altman, Anadkat et~al.}]{achiam2023gpt}
Josh Achiam, Steven Adler, Sandhini Agarwal, Lama Ahmad, Ilge Akkaya, Florencia~Leoni Aleman, Diogo Almeida, Janko Altenschmidt, Sam Altman, Shyamal Anadkat, et~al. 2023.
\newblock Gpt-4 technical report.
\newblock \emph{arXiv preprint arXiv:2303.08774}.

\bibitem[{Barak(2005)}]{barak2005purposive}
Aharon Barak. 2005.
\newblock Purposive interpretation in law.

\bibitem[{Bhat(2019)}]{bhat2019idea}
P~Ishwara Bhat. 2019.
\newblock \emph{Idea and methods of legal research}.
\newblock Oxford University Press.

\bibitem[{Coan and Surden(2024)}]{coan2024artificial}
Andrew Coan and Harry Surden. 2024.
\newblock Artificial intelligence and constitutional interpretation.
\newblock \emph{Arizona Legal Studies Discussion Paper}, (24-30).

\bibitem[{Cui et~al.(2024)Cui, Ning, Li, Chen, Yan, Li, Ling, Tian, and Yuan}]{cui2024chatlaw}
Jiaxi Cui, Munan Ning, Zongjian Li, Bohua Chen, Yang Yan, Hao Li, Bin Ling, Yonghong Tian, and Li~Yuan. 2024.
\newblock \href {https://arxiv.org/abs/2306.16092} {Chatlaw: A multi-agent collaborative legal assistant with knowledge graph enhanced mixture-of-experts large language model}.
\newblock \emph{Preprint}, arXiv:2306.16092.

\bibitem[{Dworkin(1982)}]{dworkin1982law}
Ronald Dworkin. 1982.
\newblock Law as interpretation.
\newblock \emph{Critical Inquiry}, 9(1):179--200.

\bibitem[{Endicott(2000)}]{endicott2000vagueness}
Timothy~AO Endicott. 2000.
\newblock \emph{Vagueness in law}.
\newblock Oxford University Press.

\bibitem[{Engel and Kruse(2024)}]{engel2024professor}
Christoph Engel and Johannes Kruse. 2024.
\newblock Professor gpt: Having a large language model write a commentary on freedom of assembly.

\bibitem[{Engel and McAdams(2024)}]{engel2024asking}
Christoph Engel and Richard~H McAdams. 2024.
\newblock Asking gpt for the ordinary meaning of statutory terms.
\newblock \emph{U. Ill. JL Tech. \& Pol'y}, page 235.

\bibitem[{Farnsworth et~al.(2011)Farnsworth, Guzior, and Malani}]{Farnsworth2011ImplicitBI}
W.~W. Farnsworth, Dustin~F. Guzior, and Anup Malani. 2011.
\newblock \href {https://api.semanticscholar.org/CorpusID:142783728} {Implicit bias in legal interpretation}.

\bibitem[{Fei et~al.(2025)Fei, Zhang, Shen, Zhu, Wang, Ge, and Ng}]{fei2025internlm}
Zhiwei Fei, Songyang Zhang, Xiaoyu Shen, Dawei Zhu, Xiao Wang, Jidong Ge, and Vincent Ng. 2025.
\newblock Internlm-law: An open-sourced chinese legal large language model.
\newblock In \emph{Proceedings of the 31st International Conference on Computational Linguistics}, pages 9376--9392.

\bibitem[{GLM et~al.(2024)GLM, Zeng, Xu, Wang, Zhang, Yin, Rojas, Feng, Zhao, Lai, Yu, Wang, Sun, Zhang, Cheng, Gui, Tang, Zhang, Li, Zhao, Wu, Zhong, Liu, Huang, Zhang, Zheng, Lu, Duan, Zhang, Cao, Yang, Tam, Zhao, Liu, Xia, Zhang, Gu, Lv, Liu, Liu, Yang, Song, Zhang, An, Xu, Niu, Yang, Li, Bai, Dong, Qi, Wang, Yang, Du, Hou, and Wang}]{glm2024chatglm}
Team GLM, Aohan Zeng, Bin Xu, Bowen Wang, Chenhui Zhang, Da~Yin, Diego Rojas, Guanyu Feng, Hanlin Zhao, Hanyu Lai, Hao Yu, Hongning Wang, Jiadai Sun, Jiajie Zhang, Jiale Cheng, Jiayi Gui, Jie Tang, Jing Zhang, Juanzi Li, Lei Zhao, Lindong Wu, Lucen Zhong, Mingdao Liu, Minlie Huang, Peng Zhang, Qinkai Zheng, Rui Lu, Shuaiqi Duan, Shudan Zhang, Shulin Cao, Shuxun Yang, Weng~Lam Tam, Wenyi Zhao, Xiao Liu, Xiao Xia, Xiaohan Zhang, Xiaotao Gu, Xin Lv, Xinghan Liu, Xinyi Liu, Xinyue Yang, Xixuan Song, Xunkai Zhang, Yifan An, Yifan Xu, Yilin Niu, Yuantao Yang, Yueyan Li, Yushi Bai, Yuxiao Dong, Zehan Qi, Zhaoyu Wang, Zhen Yang, Zhengxiao Du, Zhenyu Hou, and Zihan Wang. 2024.
\newblock \href {https://arxiv.org/abs/2406.12793} {Chatglm: A family of large language models from glm-130b to glm-4 all tools}.
\newblock \emph{Preprint}, arXiv:2406.12793.

\bibitem[{Guu et~al.(2020)Guu, Lee, Tung, Pasupat, and Chang}]{guu2020retrieval}
Kelvin Guu, Kenton Lee, Zora Tung, Panupong Pasupat, and Mingwei Chang. 2020.
\newblock Retrieval augmented language model pre-training.
\newblock In \emph{International conference on machine learning}, pages 3929--3938. PMLR.

\bibitem[{Hart and Green(2012)}]{hart2012concept}
Herbert Lionel~Adolphus Hart and Leslie Green. 2012.
\newblock \emph{The concept of law}.
\newblock oxford university press.

\bibitem[{Hoffman and Arbel(2024)}]{hoffman2024generative}
David~A Hoffman and Yonathan Arbel. 2024.
\newblock Generative interpretation.
\newblock \emph{New York University Law Review}, page 451.

\bibitem[{Holmes(1898)}]{holmes1898theory}
Oliver~Wendell Holmes. 1898.
\newblock Theory of legal interpretation.
\newblock \emph{Harv. L. Rev.}, 12:417.

\bibitem[{Huang et~al.(2023)Huang, Tao, Zhang, An, Jiang, Chen, Wu, and Feng}]{huang2023lawyer}
Quzhe Huang, Mingxu Tao, Chen Zhang, Zhenwei An, Cong Jiang, Zhibin Chen, Zirui Wu, and Yansong Feng. 2023.
\newblock Lawyer llama technical report.
\newblock \emph{arXiv preprint arXiv:2305.15062}.

\bibitem[{Hutchinson and Duncan(2012)}]{hutchinson2012defining}
Terry Hutchinson and Nigel Duncan. 2012.
\newblock Defining and describing what we do: doctrinal legal research.
\newblock \emph{Deakin law review}, 17(1):83--119.

\bibitem[{Jiang et~al.(2024)Jiang, Zhang, Mahari, Kessler, Ma, August, Li, Pentland, Kim, Roy, and Kabbara}]{jiang-etal-2024-leveraging}
Hang Jiang, Xiajie Zhang, Robert Mahari, Daniel Kessler, Eric Ma, Tal August, Irene Li, Alex Pentland, Yoon Kim, Deb Roy, and Jad Kabbara. 2024.
\newblock \href {https://doi.org/10.18653/v1/2024.acl-long.388} {Leveraging large language models for learning complex legal concepts through storytelling}.
\newblock In \emph{Proceedings of the 62nd Annual Meeting of the Association for Computational Linguistics (Volume 1: Long Papers)}, pages 7194--7219, Bangkok, Thailand. Association for Computational Linguistics.

\bibitem[{Kojima et~al.(2022)Kojima, Gu, Reid, Matsuo, and Iwasawa}]{kojima2022}
Takeshi Kojima, Shixiang~Shane Gu, Machel Reid, Yutaka Matsuo, and Yusuke Iwasawa. 2022.
\newblock Large language models are zero-shot reasoners.
\newblock \emph{Advances in neural information processing systems}, 35:22199--22213.

\bibitem[{Lewis et~al.(2020)Lewis, Perez, Piktus, Petroni, Karpukhin, Goyal, K{\"u}ttler, Lewis, Yih, Rockt{\"a}schel et~al.}]{lewis2020retrieval}
Patrick Lewis, Ethan Perez, Aleksandra Piktus, Fabio Petroni, Vladimir Karpukhin, Naman Goyal, Heinrich K{\"u}ttler, Mike Lewis, Wen-tau Yih, Tim Rockt{\"a}schel, et~al. 2020.
\newblock Retrieval-augmented generation for knowledge-intensive nlp tasks.
\newblock \emph{Advances in Neural Information Processing Systems}, 33:9459--9474.

\bibitem[{Liu and Li(2024)}]{liu2024judges}
John~Zhuang Liu and Xueyao Li. 2024.
\newblock How do judges use large language models? evidence from shenzhen.
\newblock \emph{Journal of Legal Analysis}, 16(1):235--262.

\bibitem[{Majeed et~al.(2023)Majeed, Hilal, Khan et~al.}]{majeed2023doctrinal}
Nasir Majeed, Amjad Hilal, Arshad~Nawaz Khan, et~al. 2023.
\newblock Doctrinal research in law: Meaning, scope and methodology.
\newblock \emph{Bulletin of Business and Economics (BBE)}, 12(4):559--563.

\bibitem[{Nyarko and Sanga(2022)}]{nyarko2022statistical}
Julian Nyarko and Sarath Sanga. 2022.
\newblock A statistical test for legal interpretation: Theory and applications.
\newblock \emph{The Journal of Law, Economics, and Organization}, 38(2):539--569.

\bibitem[{Paquin et~al.(1991)Paquin, Blanchard, and Thomasset}]{paquin1991loge}
Louis-Claude Paquin, Fran{\c{c}}ois Blanchard, and Claude Thomasset. 1991.
\newblock Loge--expert: from a legal expert system to an information system for non-lawyers.
\newblock In \emph{Proceedings of the 3rd international conference on Artificial intelligence and law}, pages 254--259.

\bibitem[{Pound(1913)}]{pound1913end}
Roscoe Pound. 1913.
\newblock The end of law as developed in legal rules and doctrines.
\newblock \emph{Harv. L. Rev.}, 27:195.

\bibitem[{{Qwen Team}(2024)}]{qwen2.5}
{Qwen Team}. 2024.
\newblock \href {https://qwenlm.github.io/blog/qwen2.5/} {Qwen2.5: A party of foundation models}.

\bibitem[{{\v{S}}avelka and Ashley(2021{\natexlab{a}})}]{savelka2021discovering}
Jarom{\'\i}r {\v{S}}avelka and Kevin~D Ashley. 2021{\natexlab{a}}.
\newblock Discovering explanatory sentences in legal case decisions using pre-trained language models.
\newblock In \emph{Findings of the Association for Computational Linguistics: EMNLP 2021}, pages 4273--4283.

\bibitem[{{\v{S}}avelka and Ashley(2021{\natexlab{b}})}]{savelka2021legal}
Jarom{\'\i}r {\v{S}}avelka and Kevin~D Ashley. 2021{\natexlab{b}}.
\newblock Legal information retrieval for understanding statutory terms.
\newblock \emph{Artificial Intelligence and Law}, pages 1--45.

\bibitem[{Savelka et~al.(2023)Savelka, Ashley, Gray, Westermann, and Xu}]{savelka2023explaining}
Jaromir Savelka, Kevin~D Ashley, Morgan~A Gray, Hannes Westermann, and Huihui Xu. 2023.
\newblock Explaining legal concepts with augmented large language models (gpt-4).
\newblock \emph{arXiv preprint arXiv:2306.09525}.

\bibitem[{Smits(2017)}]{smits2017legal}
Jan~M Smits. 2017.
\newblock What is legal doctrine? on the aims and methods of legal-dogmatic research.

\bibitem[{Tiller and Cross(2006)}]{tiller2006legal}
Emerson~H Tiller and Frank~B Cross. 2006.
\newblock What is legal doctrine.
\newblock \emph{Nw. UL Rev.}, 100:517.

\bibitem[{Van~Hoecke(2011)}]{van2011methodologies}
Mark Van~Hoecke. 2011.
\newblock Methodologies of legal research.

\bibitem[{Van~Hoecke and Ost(1998)}]{van1998legal}
Mark Van~Hoecke and Fran{\c{c}}ois Ost. 1998.
\newblock Legal doctrine in crisis: towards a european legal science.
\newblock \emph{Legal Studies}, 18(2):197--215.

\bibitem[{VanGestel and Micklitz(2011)}]{vangestel2011revitalizing}
Rob VanGestel and Hans-W Micklitz. 2011.
\newblock Revitalizing doctrinal legal research in europe: What about methodology?

\bibitem[{Waterman and Peterson(1981)}]{waterman1981models}
DA~Waterman and MA~Peterson. 1981.
\newblock Models of legal decision-making, r-2717-icj.

\bibitem[{Yue et~al.(2023)Yue, Chen, Wang, Li, Shen, Liu, Zhou, Xiao, Yun, Huang et~al.}]{yue2023disc}
Shengbin Yue, Wei Chen, Siyuan Wang, Bingxuan Li, Chenchen Shen, Shujun Liu, Yuxuan Zhou, Yao Xiao, Song Yun, Xuanjing Huang, et~al. 2023.
\newblock Disc-lawllm: Fine-tuning large language models for intelligent legal services.
\newblock \emph{arXiv preprint arXiv:2309.11325}.

\bibitem[{Yung-chin Su(2024)}]{Su2024}
Zhang~Cheng Yung-chin Su, Zhou~Xiang. 2024.
\newblock The future of the legal dogmatics in the perspective of structure and management:new frontier of theoretical dogmatics with practical dogmatics in ai's hand (in chinese).
\newblock \emph{NanJing University Law Journal}, 1:1--17.

\bibitem[{Zheng et~al.(2023)Zheng, Chiang, Sheng, Zhuang, Wu, Zhuang, Lin, Li, Li, Xing et~al.}]{zheng2023judging}
Lianmin Zheng, Wei-Lin Chiang, Ying Sheng, Siyuan Zhuang, Zhanghao Wu, Yonghao Zhuang, Zi~Lin, Zhuohan Li, Dacheng Li, Eric Xing, et~al. 2023.
\newblock Judging llm-as-a-judge with mt-bench and chatbot arena.
\newblock \emph{Advances in Neural Information Processing Systems}, 36:46595--46623.

\bibitem[{Zhu et~al.(2023)Zhu, Wang, and Wang}]{zhu2023judgelm}
Lianghui Zhu, Xinggang Wang, and Xinlong Wang. 2023.
\newblock Judgelm: Fine-tuned large language models are scalable judges.
\newblock \emph{arXiv preprint arXiv:2310.17631}.

\end{thebibliography}

\onecolumn
\appendix

\clearpage
\section{The structure of cases}
\label{judgment_introduction}
A case in China can generally be divided into five sections: header, facts, court view, verdict, and conclusion. The \textbf{header} includes the name of the court, the type of document, case number, basic information about the parties involved, the origin of the case, and details about the judicial panel and trial method. The \textbf{facts} section outlines the plaintiff's claims, facts, arguments, and the defendant's admissions regarding the plaintiff's factual assertions. The \textbf{court view} section provides the rationale for the judgment and the legal basis upon which it is made. The \textbf{verdict} contains the decision on substantive issues of the case. Finally, the \textbf{conclusion} ends the judgment document formally.

\section{Examples of relevant cases}
\label{appendix:relevant_case}
\begin{table*}[h]
  \centering
   \scalebox{0.8}{

   \begin{tabular}{|p{2.5cm}|p{2.3cm}|p{5cm}|p{8.2cm}|}
    \hline
    \textbf{Charge} & \textbf{Vague concept} & \textbf{Cases \textit{mentioning} the concept (Irrelevant Cases)} & \textbf{Cases that \textit{analyze} the concept in detail (Relevant Cases)} \\ \hline
    Theft & Dwelling & The court holds that the defendant, Yang, with the intent of unlawful possession, secretly entered a \textit{dwelling} to steal another person's property. His actions constitute the crime of theft...
    & Regarding whether Zhang's actions constitute theft by entering a \textit{dwelling}, upon investigation, location A is an employee dormitory rented by B restaurant. Although it is relatively isolated from the outside, it lacks clear features of serving household living functions and should not be recognized as entering a \textit{dwelling}.
    \\ \hline
    Traffic accident crime & Flee the scene & After the accident, the defendant \textit{fled the scene} and is fully responsible for the incident. His actions constitute the crime of traffic accident liability as stipulated in Article 133 of the Criminal Law of the People's Republic of China. & The defendant argues that after the accident, he had his wife promptly dial 120 for emergency assistance and then left the scene to return home, claiming that he did not flee. Upon investigation, it is confirmed that the defendant did call 120 in a timely manner, but this action was not reported to the authorities. After learning that the victim had died, the defendant fled the scene. His actions should be recognized as \textit{fleeing}, and his defense is not accepted.
    \\ \hline
    
    \end{tabular}
}
  \caption{Cases mentioning the vague concept and Cases discussing in detail why the vague concept applies. We only consider the latter to be the relevant cases.}

  \label{tab:case}
\end{table*}

\section{Details of ATRIE}
\label{eval_details}

\subsection{Manual inspection of the LLM-annotated data}
\label{appendix:manual_inspection}
To evaluate the relevance between the LLM-filtered cases and the vague concepts, we randomly sampled 20 cases for each concept from $\mathcal{D}$ and manually assessed their relevance to the vague concepts. The results show that over 96\% of the cases are indeed relevant to the vague concepts. In addition, manual inspection of 200 extraction results indicates that the accuracy of Qwen2.5-72B-Instruct in labeling the gold label $l$ and the reasoning $r$ are 98\% and 94\%, respectively.

\subsection{Example of gold labels and reasons}
\label{appendix:label}
Table \ref{tab:data_example} shows some examples of gold labels and reasons in the LCE dataset.
\begin{table}[h]
  \centering
   \scalebox{1}{

   \begin{tabular}{p{1cm}p{13cm}}
    \toprule
    Label & Reason \\ \midrule
    Yes & The location of the theft is a closed store that integrates living quarters and business operations. Since the store is connected to the living area, and after closing, it becomes part of the living space, relatively isolated from the outside, this theft is classified as theft by entering a dwelling.
\\ \hline
    No & The dormitory is a collective dormitory of the factory, intended solely for employees to rest during lunch breaks and nighttime. It does not include facilities for dining or other living functions and lacks the characteristics of a dwelling. Therefore, the accusation of the defendant committing theft by entering a dwelling is inappropriate.
 \\ \bottomrule
    
    \end{tabular}
}
  \caption{Examples of gold labels and their corresponding gold reasons
.}

  \label{tab:data_example}
\end{table}

\subsection{Detailed results}
\label{appendix:results}
\subsubsection{Different models} 
\label{appendix:different_models}
As shown in Table~\ref{tab:diffmodel}, to validate the generalizability of our method, we utilized different LLMs to generate interpretations and perform automated evaluations. Due to the cost constraints of APIs, we conducted experiments on a subset of our LCE dataset. Our findings are as follows: 
(1) \textbf{Stronger models demonstrate more remarkable ability to generate concept interpretations.
}The interpretations generated using Qwen2.5 (72B) and GPT-4o lead to noticeably higher performance improvements 
than using GPT-4o-mini.
(2) \textbf{Generated concept interpretations can assist even weaker LLMs in accurately understanding vague concepts.}
In our method, the performance gap between GLM and the other models is significantly smaller than that observed in the Zero-Shot baseline.

\begin{table*}[h]
  \centering
   \scalebox{1}{
  \begin{tabular}{c|ccc|ccc|ccc} 
     \hline
    Interpret model& \multicolumn{3}{c}{Qwen2.5 (72B)}
& \multicolumn{3}{c}{gpt-4o-2024-08-06} 
&  \multicolumn{3}{c}{gpt-4o-mini} 
\\ 
 \cline{1-10} Predict model
 & Qwen & GPT& GLM
& Qwen & GPT& GLM
& Qwen & GPT& GLM
\\ 
 \hline
    
    Zero-Shot
    & 57.27 & 51.68 & 47.06
    & 57.27 & 51.68 & 47.06
    & 57.27 & 51.68 & 47.06\\ 
    \hline
    Direct Interpretation
    & 61.58 & 53.65 & 53.14
    & 61.02 & 52.70 & 54.96
    & 55.94 & 51.80 & 50.15\\ 
    Judicial Interpretation
    & 62.14 & \textbf{59.05}& 53.05
    & \textbf{62.14} & 59.05& 53.05
    & 62.14 & \textbf{59.05}& 53.05\\ 
    ATRIE
    & \textbf{66.67} & 59.01 & \textbf{60.34}
    & 61.99 & \textbf{60.01} & \textbf{59.23}
    & \textbf{63.14} & 54.14 & \textbf{54.18}\\ 
   
 \hline
  \end{tabular}
  }
  \caption{\label{tab:diffmodel}
    Macro-F1 results of using different LLMs to generate interpretations and perform the Legal Concept Entailment task on a subset. Here, \textbf{Qwen}, \textbf{GPT}, and \textbf{GLM} represent Qwen2.5-72B-Instruct, gpt-4o-mini, and GLM-4-9B-Chat\cite{glm2024chatglm}, respectively.
  }

\end{table*}

\subsubsection{Model bias}
Analyzing the LLM's predictions reveals a strong bias toward responding with "Yes" on the Legal Concept Entailment task (Table~\ref{tab:pred_label}). 
This is one of the reasons we perform label balancing on the LCE dataset. If the dataset consists solely of positive examples, it becomes challenging to effectively evaluate the LLM's performance on the LCE task.
\begin{table}[h]
  \centering
   \scalebox{1}{ 

  \begin{tabular}{ccccccc}
    \hline
    &  \multicolumn{3}{c}{Qwen2.5 (72B)}&\multicolumn{3}{c}{Qwen2.5 (14B)} \\
    & Pos& Neg & Ratio & Pos  & Neg & Ratio\\
    \hline

    Zero-Shot           & 2285 & 367 & 6.23 & 2329 & 323 & 7.21 \\
    Chain-of Thought          & 2216 & 436 & 5.08 & 2313 & 338 & 6.84 \\
    Direct Interpretation           & 1989 & 662 & 3.00 & 2049 & 602 & 3.40 \\
    Judicial Interpretation           & 2018 & 634 & 3.18 & 2011 & 641 & 3.14 \\
    ATRIE         & 1939 & 713 & 2.72 & 1926 & 726 & 2.65 \\
    \hline
    Gold Label         & 1714 & 837 & 2.05 & 1714 & 837 & 2.05 \\
    \hline
  \end{tabular}
}
  \caption{The number and ratio of positive and negative cases predicted by the LLM. \textit{Pos} represents the number of cases predicted as "Yes", \textit{Neg} represents the number of cases predicted as "No", and \textit{Ratio} denotes the ratio of \textit{Pos} to \textit{Neg}.}
  \label{tab:pred_label}
\end{table}

\subsection{\revise{Open-source LLMs are also good evaluators}}
\label{appendix:llm_cs}
The primary objective of using LLMs as evaluators in our work is to assess the consistency between the reasoning processes of LLM outputs and the reference answers. In our main experiments, we use GPT-
4o as the evaluator, but open-source LLMs can also effectively evaluate this consistency. We compared evaluation results in Table~\ref{tab:consistency}, finding that the Spearman correlation coefficients between GPT-4o and Qwen2.5 (72B)/Qwen2.5 (32B) scores are 0.943 and 0.829, respectively. This demonstrates that using the open-source Qwen2.5 (72B) for evaluation yields results comparable to GPT-4o. 
\begin{table}[h]
    \centering
    \scalebox{0.9}{
    \begin{tabular}{l|cc|cc|cc}
        \hline
        & \multicolumn{2}{c|}{GPT-4o} & \multicolumn{2}{c|}{Qwen2.5 (72B)} & \multicolumn{2}{c}{Qwen2.5 (32B)} \\
         & CS & Ranking & CS & Ranking & CS & Ranking \\
        \hline
        Zero-Shot & 5.658 & 3 & 5.481 & 4 & 5.589 & 5 \\
        Chain-of-Thought & 5.717 & 2 & 5.764 & 2 & 5.856 & 2 \\
        Judicial Interpretation & 5.573 & 6 & 5.425 & 6 & 5.562 & 6 \\
        Expert Interpretation & 5.630 & 5 & 5.456 & 5 & 5.642 & 4 \\
        Direct Interpretation & 5.642 & 4 & 5.599 & 3 & 5.753 & 3 \\
        ATRIE & 5.946 & 1 & 5.848 & 1 & 6.006 & 1 \\
        \hline
    \end{tabular}
    }
    \caption{Evaluation results of different LLMs on consistency between the reasoning processes of LLM outputs and reference answers.}
    \label{tab:consistency}
\end{table}
\subsection{\revise{Why don't we use legal LLMs in our interpreter?}}
\label{app:legalllm}
We considered utilizing more Chinese legal LLMs apart from Farui for generating concept interpretations. However, since this task requires analyzing a large number of cases simultaneously, and legal LLMs lack long-text reasoning capabilities, their performance on this task was not as good as that of general-purpose LLMs. Furthermore, general-purpose LLMs currently perform very well in legal domain benchmarks, with few gaps compared to legal-specific LLMs. Considering these two points, we ultimately decide to only use general-purpose LLMs in our main experiments. 
\paragraph{The context length of existing legal LLMs cannot meet the task requirements}
Our task requires summarizing vague concept interpretations from a large number of cases, necessitating that the LLM can analyze many cases simultaneously. The average length of relevant text extracted from a single case is 96 tokens. In our experiments, we typically need to analyze 166 cases simultaneously, resulting in an average input length of 17k tokens per concept. Table~\ref{tab:legalllm} lists most existing Chinese legal LLMs, their availability, and their context lengths. From the table, we can see that the current Chinese legal LLMs either are not available for use, such as InternLM-Law and ChatLaw2-MoE, or have insufficient context lengths, such as DISC-LawLLM and ChatLaw-33B. Farui-plus has a relatively longer context length among the usable legal LLMs, so we selected it for experiments. 

We control the input length within 10k tokens and compare the concept interpretation generated by farui-plus and Qwen2.5-72B. Table~\ref{tab:legal_llm} shows the results.
Although Farui-plus claims an input length of up to 12k, we find in practice that when the output length exceeds 5k, its instruction-following ability is significantly weaker than that of general-purpose LLMs, and it even fails to produce outputs in the expected format and content.
\begin{table}[ht]
\centering
\begin{tabular}{|l|l|l|}
\hline
\textbf{Model}          & \textbf{Availability} & \textbf{Max Context Length} \\ \hline
InternLM-Law~\cite{fei2025internlm}         & No                    & $\geq$ 32k                    \\ \hline
ChatLaw2-MoE~\cite{cui2024chatlaw}         & No                    & Unknown                       \\ \hline
Farui-plus           & Yes                   & 12k                           \\ \hline
DISC-LawLLM~\cite{yue2023disc}          & Yes                   & 4096                          \\ \hline
ChatLaw-33B~\cite{cui2024chatlaw}          & Yes                   & 2048                          \\ \hline
Lawyer LLaMA~\cite{huang2023lawyer}         & Yes                   & 2048                          \\ \hline
\end{tabular}
\caption{Availability and Max Context Length of Chinese legal LLMs}
\label{tab:legalllm}
\end{table}

\paragraph{General-purpose LLMs perform well on legal tasks}
General-purpose LLMs possess sufficient legal knowledge and reasoning abilities. As evidenced by~\citet{fei2025internlm}, Qwen1.5-72B achieves the best performance on LawBench, except for the unreleased InternLM-Law-7B, even surpassing GPT-4. We reasonably infer that its upgraded version, Qwen2.5-72B, can also offer sufficient legal reasoning capacity, since it outperforms Qwen1.5 versions by a large margin across various benchmarks.
We thus use strong general-purpose LLMs with long-context reasoning abilities in our experiments. We will investigate this issue  again when proper legal LLMs with such capacities become available.

\section{\revise{The efficiency of our framework}}
\label{app:efficency}
Our framework provides a cost-effective solution for legal concept interpretation tasks, significantly reducing reliance on senior legal experts. For one concept, our framework only requires 3.6 A40 GPU hours to filter 13k cases and find 332 useful cases, costing only 1.5 dollars. We also recruit two legal experts who had passed China's Unified Qualification Exam for Legal Professionals, instructing them to independently write 5 concept interpretations in total based solely on court judgments, legal textbooks, and other materials without referencing existing concept interpretations. The average time spent on manually crafting each concept interpretation is 2 hours, but they only analyze less than 50 cases.
The cost of hiring legal experts to draft a concept interpretation is 20 dollars.
Our framework demonstrates remarkable efficiency by enabling the reading and summarization of significantly more cases while requiring substantially less time and financial investment.

\section{Details about human evaluation}
\label{appendix:human_eval}
\subsection{Details about evaluation metrics}
\begin{itemize}
    \item \textbf{Accuracy (Acc.)} The interpretation should align with the current legal articles and relevant judicial interpretations, avoiding any misinterpretation or distortion of the original intent of the law.
    
    \item \textbf{Informativeness (Info.)} The interpretation should provide additional previously unknown insights, thereby enhancing the human evaluators' legal knowledge beyond their prior understanding.
    
    \item \textbf{Normativity (Norm.)} The interpretation should conform to the standard expressions and terminology used in legal studies.
    
    \item \textbf{Comprehensiveness (Comp.)} The interpretation should cover as many relevant scenarios as possible, including applicable and excluded cases, ensuring no key aspects are omitted.
    
    \item \textbf{Readability (Read.)} The interpretation should be expressed in clear, simple language, avoiding excessive legal jargon or complex sentence structures so that even non-experts can generally understand the meaning and application of the legal concept.
\end{itemize}

\subsection{\revise{Instructions given to annotators}}
We shuffled the concept interpretations from different sources to ensure that annotators could evaluate each interpretation fairly and objectively. They were required not to discuss and to score independently.
The annotators achieved moderate inter-annotator agreement (Spearman's $\rho=0.42 $), with the average evaluation scores presented in Table~\ref{tab:human_eval} in our paper.

\subsection{Case Study}
\label{appendix:human_eval_case_study}
We asked legal experts to conduct a detailed analysis comparing the Direct Interpretation (DI) and our framework ATRIE, which can demonstrate how high-scoring interpretations are concretely superior to low-scoring ones. They both interpret the concept of “serious circumstances” under Article 359 of the Chinese Criminal Law. We believe the difference in scores is a reliable measure of the quality of the interpretations. 
\begin{itemize}
    \item For Accuracy (8.0 vs. 6.5), ATRIE gives a more legally faithful and analytically precise explanation by explicitly integrating quantifiable benchmarks such as the number of individuals involved and the amount of illegal profit. In contrast, DI does not give detailed criteria, reducing its alignment with actual judicial practice.
    \item In Informativeness (7.5 vs. 6.5), ATRIE summarizes a more comprehensive account of various scenarios, incorporating factors such as the means of committing the offense and the duration of the behavior. While they are partially addressed or entirely absent in DI.
    \item For Normativity (8.0 vs. 7.0), ATRIE maintains consistent use of legal language and adheres more strictly to legal theory, while DI exhibits a more colloquial tone and lacks binding legal texts, weakening its formal authority.
    \item For Comprehensiveness (8.0 vs. 6.5), ATRIE provides a fuller framework by offering both positive and negative case illustrations derived from actual judicial decisions, each structured with clear factual context and legal reasoning. This contrasts with DI's more abstract and potentially fictional examples.
    \item For the hallucination problem, We would like to thank the reviewer for pointing it out, which is really critical in the legal domain. In the case study, we manually verified all the cited cases in ATRIE. All cases are from actual judicial decisions and could be referred to specific sentences in judgment documents. There are no instances of hallucinated or fabricated content. In contrast, the examples in DI appear to be hypothetical or generalized, too abstract to be identified in cases, which adds the risk of hallucination.
\end{itemize}

\subsubsection{Direct Interpretation}
\begin{verbatim}
在中华人民共和国刑法第三百五十九条中，"引诱、容留、介绍他人卖淫的，处五年以下有期徒刑、拘役或者管制，并处罚金；情节严重的，处五年以上有期徒刑，并处罚金。" 该条文的实施中，其中的“情节严重”可能会产生一定的法律解释上的模糊性。司法程序中，法官需要根据案件的具体情况对“情节严重”的定义进行具体化和解释。

### 解析

1. **基本定义**：
    - 根据最高人民法院、最高人民检察院《关于办理组织、强迫、引诱、容留、介绍卖淫刑事案件适用法律若干问题的解释》，“情节严重”通常指行为的社会危害性较大，如引诱、容留、介绍卖淫的人数较多、次数频繁、造成严重后果等。
    - “情节严重”不仅包括行为的数量和频率，还包括行为的性质、手段、后果以及行为人的主观恶性等因素。

2. **具体适用**：
    - 对于“情节严重”进行具体适用时，需要综合考虑行为的次数、人数、手段、后果等多方面因素。
    - 例如，引诱、容留、介绍未成年人、孕妇、智障人员等特殊群体卖淫，或者使用暴力、威胁等手段，通常会被认定为“情节严重”。
    - 在具体案件中，法官会根据行为人的行为方式、行为对象、行为后果等进行综合判断。

3. **排除情况**：
    - 行为次数较少、人数较少、未造成严重后果的。
    - 行为手段较为温和，未使用暴力、威胁等手段。
    - 行为对象为一般成年人，且未涉及特殊群体。
    - 行为人的主观恶性较小，如初次犯罪、因生活所迫等。
    - 行为未对社会造成较大负面影响。

### 举例说明

- **符合“情节严重”定义的案例**：

  1. **案例一**：
     - **具体情形**：被告人多次引诱多名未成年人卖淫。
     - **案例说明**：被告人王某多次通过网络平台引诱多名未成年人卖淫，严重侵犯了未成年人的身心健康。
     - **判决结果**：法院认定其行为“情节严重”，判处五年以上有期徒刑，并处罚金。

  2. **案例二**：
     - **具体情形**：被告人使用暴力手段强迫他人卖淫。
     - **案例说明**：被告人李某使用暴力手段强迫多名女性卖淫，造成多名女性身体和心理严重伤害。
     - **判决结果**：法院认定其行为“情节严重”，判处五年以上有期徒刑，并处罚金。

  3. **案例三**：
     - **具体情形**：被告人组织卖淫活动，涉及人数众多，持续时间长。
     - **案例说明**：被告人张某组织卖淫活动长达两年，涉及数十名女性，严重扰乱了社会秩序。
     - **判决结果**：法院认定其行为“情节严重”，判处五年以上有期徒刑，并处罚金。

  4. **案例四**：
     - **具体情形**：被告人引诱、容留、介绍智障人员卖淫。
     - **案例说明**：被告人赵某多次引诱、容留、介绍智障人员卖淫，严重侵犯了智障人员的权益。
     - **判决结果**：法院认定其行为“情节严重”，判处五年以上有期徒刑，并处罚金。

  5. **案例五**：
     - **具体情形**：被告人引诱、容留、介绍卖淫活动，导致被害人自杀。
     - **案例说明**：被告人刘某多次引诱、容留、介绍他人卖淫，导致一名被害人因无法承受心理压力而自杀。
     - **判决结果**：法院认定其行为“情节严重”，判处五年以上有期徒刑，并处罚金。

- **不符合“情节严重”定义的案例**：

  1. **案例一**：
     - **具体情形**：被告人初次引诱一名成年人卖淫，未造成严重后果。
     - **案例说明**：被告人陈某因生活所迫，初次引诱一名成年人卖淫，未造成严重后果。
     - **判决结果**：法院认定其行为不属于“情节严重”，判处五年以下有期徒刑，并处罚金。

  2. **案例二**：
     - **具体情形**：被告人引诱、容留、介绍卖淫活动，但仅涉及一两名成年人。
     - **案例说明**：被告人林某引诱、容留、介绍两名成年人卖淫，未使用暴力手段，未造成严重后果。
     - **判决结果**：法院认定其行为不属于“情节严重”，判处五年以下有期徒刑，并处罚金。

  3. **案例三**：
     - **具体情形**：被告人因受他人胁迫而引诱、容留、介绍他人卖淫。
     - **案例说明**：被告人杨某因受他人胁迫，被迫引诱、容留、介绍他人卖淫，主观恶性较小。
     - **判决结果**：法院认定其行为不属于“情节严重”，判处五年以下有期徒刑，并处罚金。

  4. **案例四**：
     - **具体情形**：被告人引诱、容留、介绍卖淫活动，但未造成被害人身体或心理严重伤害。
     - **案例说明**：被告人周某引诱、容留、介绍他人卖淫，但未使用暴力手段，且被害人未受到严重伤害。
     - **判决结果**：法院认定其行为不属于“情节严重”，判处五年以下有期徒刑，并处罚金。

  5. **案例五**：
     - **具体情形**：被告人引诱、容留、介绍卖淫活动，但行为时间较短，涉及人数较少。
     - **案例说明**：被告人黄某在短时间内引诱、容留、介绍三名成年人卖淫，未造成严重后果。
     - **判决结果**：法院认定其行为不属于“情节严重”，判处五年以下有期徒刑，并处罚金。

### 司法裁量

法官在具体案件中对“情节严重”的定义的司法裁量主要基于以下几点：
- **行为次数**：行为的次数越多，越容易被认定为“情节严重”。
- **行为人数**：涉及的人数越多，越容易被认定为“情节严重”。
- **行为手段**：使用暴力、威胁等手段，或者引诱特殊群体（如未成年人、智障人员）卖淫，通常会被认定为“情节严重”。
- **行为后果**：行为造成严重后果，如被害人身体或心理严重伤害、社会秩序严重扰乱等，通常会被认定为“情节严重”。
- **行为人的主观恶性**：行为人的主观恶性较大，如多次犯罪、有前科等，通常会被认定为“情节严重”。

综上，“情节严重”在独立案件中需结合行为的次数、人数、手段、后果以及行为人的主观恶性等多方面因素进行综合认定，以确保判决的法律公正。
\end{verbatim}

\subsubsection{ATRIE}
\begin{verbatim}
在中华人民共和国刑法第三百五十九条中，"引诱、容留、介绍他人卖淫的，处五年以下有期徒刑、拘役或者管制，并处罚金；情节严重的，处五年以上有期徒刑，并处罚金"涉及引诱、容留、介绍卖淫行为的定罪和量刑，其中的“情节严重”可能会产生一定的法律解释上的模糊性。司法程序中，法官需要根据案件的实际情况对“情节严重”的定义进行具体化和解释。

### 解析

1. **基本定义**：
    - 根据《最高人民法院、最高人民检察院关于办理组织、强迫、引诱、容留、介绍卖淫刑事案件适用法律若干问题的解释》第九条，情节严重的情形包括但不限于：
        - 引诱、容留、介绍十人以上卖淫的；
        - 引诱、容留、介绍五人以上未成年人、孕妇、智障人员、患有严重性病的人卖淫的；
        - 非法获利人民币五万元以上的；
        - 造成恶劣社会影响的；
        - 其他情节严重的情形。

2. **具体适用**：
    - 对于“情节严重”进行具体适用时，需要综合考虑以下几个方面：
        - **人数**：引诱、容留、介绍卖淫的人数是否达到十人以上。
        - **特定对象**：是否涉及未成年人、孕妇、智障人员、患有严重性病的人等特定对象。
        - **非法获利**：非法获利金额是否达到五万元以上。
        - **社会影响**：是否造成了恶劣的社会影响。
        - **持续时间**：行为的持续时间是否较长。
        - **犯罪手段**：是否使用了特别恶劣的手段，如暴力、威胁等。

3. **排除情况**：
    - **人数不足**：引诱、容留、介绍卖淫的人数未达到十人以上。
    - **特定对象不足**：涉及的特定对象人数未达到五人以上。
    - **非法获利不足**：非法获利金额未达到五万元以上。
    - **社会影响不大**：未造成恶劣的社会影响。
    - **时间较短**：行为的持续时间较短。
    - **手段一般**：未使用特别恶劣的手段。

### 举例说明

- **符合“情节严重”定义的案例**：

  1. **案例一**：
     - **具体情形**：被告人潘素琴、王某在其招待所内多次容留他人卖淫，次数达100余次，非法获利1000余元。
     - **案例说明**：被告人潘素琴、王某在其经营的招待所内多次容留他人卖淫，次数和非法获利均达到较高标准，且持续时间较长，法院认定其行为属于情节严重。
     - **判决结果**：法院认定其行为属于情节严重，判处五年以上有期徒刑。

  2. **案例二**：
     - **具体情形**：被告人刘军在其经营的“君安沐浴休闲会所”内长期容留他人卖淫，非法获利达10万元左右。
     - **案例说明**：被告人刘军在其经营的会所内长期容留他人卖淫，非法获利金额较高，且持续时间较长，法院认定其行为属于情节严重。
     - **判决结果**：法院认定其行为属于情节严重，判处五年以上有期徒刑。

  3. **案例三**：
     - **具体情形**：被告人陈信涵、朱萍在其经营的场所内容留、介绍12名女性卖淫，其中5名系未成年人，非法获利金额较高。
     - **案例说明**：被告人陈信涵、朱萍在其经营的场所内容留、介绍多人卖淫，且涉及多名未成年人，非法获利金额较高，法院认定其行为属于情节严重。
     - **判决结果**：法院认定其行为属于情节严重，判处五年以上有期徒刑。

  4. **案例四**：
     - **具体情形**：被告人周辉在其经营的“方圆浴场”内长期容留多名妇女卖淫，非法获利247748元。
     - **案例说明**：被告人周辉在其经营的浴场内长期容留多名妇女卖淫，非法获利金额较高，且持续时间较长，法院认定其行为属于情节严重。
     - **判决结果**：法院认定其行为属于情节严重，判处五年以上有期徒刑。

  5. **案例五**：
     - **具体情形**：被告人张福春在其经营的场所内容留10人以上卖淫。
     - **案例说明**：被告人张福春在其经营的场所内容留10人以上卖淫，符合司法解释中关于人数的规定，法院认定其行为属于情节严重。
     - **判决结果**：法院认定其行为属于情节严重，判处五年以上有期徒刑。

- **不符合“情节严重”定义的案例**：

  1. **案例一**：
     - **具体情形**：被告人龚丽平在其经营的洗头房内容留他人卖淫20人次以上，但非法获利金额不高，且未造成严重社会影响。
     - **案例说明**：被告人龚丽平虽然容留他人卖淫次数较多，但非法获利金额不高，且未造成严重社会影响，法院认定其行为不属于情节严重。
     - **判决结果**：法院认定其行为不属于情节严重，判处五年以下有期徒刑。

  2. **案例二**：
     - **具体情形**：被告人马某甲在其经营的场所内容留五人卖淫，但非法获利金额不高，且未造成严重社会影响。
     - **案例说明**：被告人马某甲虽然容留多人卖淫，但非法获利金额不高，且未造成严重社会影响，法院认定其行为不属于情节严重。
     - **判决结果**：法院认定其行为不属于情节严重，判处五年以下有期徒刑。

  3. **案例三**：
     - **具体情形**：被告人丘某娟在其出租屋内容留三人卖淫，时间较短，非法获利金额不高。
     - **案例说明**：被告人丘某娟虽然容留多人卖淫，但时间较短，非法获利金额不高，且未造成严重社会影响，法院认定其行为不属于情节严重。
     - **判决结果**：法院认定其行为不属于情节严重，判处五年以下有期徒刑。

  4. **案例四**：
     - **具体情形**：被告人胡某甲在其经营的场所内介绍四名妇女卖淫，非法获利金额不高。
     - **案例说明**：被告人胡某甲虽然介绍多人卖淫，但非法获利金额不高，且未造成严重社会影响，法院认定其行为不属于情节严重。
     - **判决结果**：法院认定其行为不属于情节严重，判处五年以下有期徒刑。

  5. **案例五**：
     - **具体情形**：被告人韦某在其经营的场所内容留三人卖淫，时间较短，非法获利金额不高。
     - **案例说明**：被告人韦某虽然容留多人卖淫，但时间较短，非法获利金额不高，且未造成严重社会影响，法院认定其行为不属于情节严重。
     - **判决结果**：法院认定其行为不属于情节严重，判处五年以下有期徒刑。

### 司法裁量

法官在具体案件中对“情节严重”的定义的司法裁量主要基于以下几点：
- **人数**：确认引诱、容留、介绍卖淫的人数是否达到十人以上。
- **特定对象**：确认是否涉及未成年人、孕妇、智障人员、患有严重性病的人等特定对象，且人数是否达到五人以上。
- **非法获利**：确认非法获利金额是否达到五万元以上。
- **社会影响**：确认是否造成了恶劣的社会影响。
- **持续时间**：确认行为的持续时间是否较长。
- **犯罪手段**：确认是否使用了特别恶劣的手段，如暴力、威胁等。

综上，“情节严重”在独立案件中需结合案件的具体情况，如人数、特定对象、非法获利、社会影响、持续时间和犯罪手段等进行综合认定，以确保判决的法律公正。
\end{verbatim}

\section{Details of the generated concept interpretation}
\label{appendix:interpretation}
\subsection{The structure of generated concept interpretation}
\label{appendix:interpretation_structure}
The generated concept interpretation includes the following main components.
This structure is finalized after being generated by LLM and modified by legal experts.

\begin{itemize} 
\item \textbf{Analysis}: Cites judicial interpretations 
 or other legal text to define the vague concept's basic meaning, applicability conditions, and exclusions. 
\item \textbf{Example Cases}: Provides specific case examples illustrating how the vague concept is applied; this section includes 5 Positive Cases and 5 Negative Cases. 
\item \textbf{Judicial Discretion}: Provides multiple judgment criteria to guide judges on how to flexibly apply the vague concept based on the case's specifics. 
\end{itemize}

\subsection{Details of the interpretation example $e_0$}
\label{appendix:e_0}
We additionally select a vague concept $c_0$ and its corresponding article $a_0$. 
$c_0$ and $a_0$ are not the same as any of the concepts and articles selected in Section \ref{dataset}.
Using the methods outlined in Section \ref{method}, we derive a reason set $\mathcal{R}_0$. These three components serve as input to the LLM. We generate multiple distinct interpretations. A legal expert selects one interpretation that best adheres to legal format specifications and modifies it to ensure correctness and clarity. We designate the revised interpretation as the interpretation example $e_0$.
\subsection{An example of generated vague concept interpretation}
\label{appendix:example_of_interpretation}
\subsubsection{Original text in Chinese}
\begin{verbatim}
在中华人民共和国刑法第二百六十四条中，“盗窃公私财物，数额较大的”涉及盗窃行为的定罪和量刑，该条文的实施中，其中的“入户盗窃”中“户”的概念可能会产生一定的法律解释上的模糊性。司法程序中，法官需要根据案件的实际情况对“户”的定义进行具体化和解释。

### 解析

1. **基本定义**：
    - 根据最高人民法院、最高人民检察院《关于办理盗窃刑事案件适用法律若干问题的解释》，“户”的特征表现为供他人家庭生活和与外界相对隔离的两个方面。
    - “户”通常包括家庭的居住场所、封闭的院落、为生活租用的房屋等。
    - 非法进入他人生活区域与外界相对隔离的住所盗窃的，应当认定为“入户盗窃”。

2. **具体适用**：
    - 对于“户”进行具体适用时，需要查看被盗场所是否符合供他人家庭生活的场所，并且与外界相对隔离。
    - 对于公共场所、商业用途的场所或者未经明确隔离的区域，一般不被认定为“户”。
    - 在具体案件中，法官会根据房屋的用途、侵入方式、时间等切实情况进行判断。

3. **排除情况**：
    - 不符合“生活用途”：如仅为商业用途的店铺、公共办公场所等。
    - 不具备“相对隔离性”：如无任何封闭、开放性极强的场所。
    - 他人同意或者空置：如经允许进入的情况下进行盗窃，或者在实际无人生活的装修或空房中进行盗窃。

### 举例说明

- **符合“户”定义的案例**：

  1. **案例一**：
     - **具体情形**：被告人非法进入供他人家庭生活的封闭住所进行盗窃行为。
     - **案例说明**：被告人余某甲非法进入xx区xxx村xxx号305室，该305室是他人租住的住宅，具有供家庭生活和与外界相对隔离的特征，符合“户”的定义。
     - **判决结果**：法院认定其为入户盗窃，因其非法进入相对隔离的私人住宅内实施盗窃。

  2. **案例二**：
     - **具体情形**：被告人多次进入他人家庭住所在家人不在场的情况下进行盗窃。
     - **案例说明**：被告人李某某的两次盗窃行为发生在被害人的住宅内，该住宅具有供家庭生活和与外界相对隔离的特征，符合“户”的定义。
     - **判决结果**：法院认定其为入户盗窃，因其非法进入供家庭生活的住所。

  3. **案例三**：
     - **具体情形**：被告人深夜翻墙进入与外界隔离的家庭院落，并进入室内实施盗窃。
     - **案例说明**：被告人田某深夜侵入多户被害人家中实施盗窃，这些住所均符合供家庭生活和与外界相对隔离的特征。
     - **判决结果**：法院认定其为入户盗窃，因其非法进入家庭生活用的封闭场所。

  4. **案例四**：
     - **具体情形**：被告人利用工具撬锁，破门进入封闭的私人住所实施盗窃。
     - **案例说明**：被告人张某某利用窃取的钥匙进入被害人黄某某家中实施盗窃，该住宅具有供家庭生活和与外界相对隔离的特征。
     - **判决结果**：法院认定其为入户盗窃，因其非法进入私人家庭住所。

  5. **案例五**：
     - **具体情形**：被告人在家人经常出入的生活区域安静时间段入内盗窃。
     - **案例说明**：被告人王某某多次采用秘密手段窃取公民财物，且其行为发生在户内，即被害人的住宅内。
     - **判决结果**：法院认定其为入户盗窃，因其非法进入供他人家庭生活且与外界相对隔离的场所。

- **不符合“户”定义的案例**：

  1. **案例一**：
     - **具体情形**：被告人盗窃商业用途的未居住店铺内的财物或者在公共区域内实施盗窃。
     - **案例说明**：被告人刘某某在被害人经营的商铺实施盗窃，而非进入被害人家庭生活的住所。
     - **判决结果**：法院认定其不属于入户盗窃，因为商铺主要用于商业经营，不符合“户”的定义。

  2. **案例二**：
     - **具体情形**：被告人在装修未居住的房屋中实施盗窃行为。
     - **案例说明**：被告人张某某盗窃的场所是出租楼一楼用于停放车辆的公共场所，不属于严格意义上的户。
     - **判决结果**：法院认定其不属于入户盗窃，因为该房屋未用于居住且不符合“户”的定义。

  3. **案例三**：
     - **具体情形**：被告人在被害人用来经营的场所内盗窃，但该场所主要功能为商业用途并不具备生活属性。
     - **案例说明**：被告人刘某某进入悬挂“中国移动通信／雨露／指定专营店”的店铺，该店位于被害人家庭所有房屋的一层，一层前部区域为手机经营区，一层后部及楼上区域为曾某家居住生活区，案发时一楼营业区与生活区被墙、门明确隔离。
     - **判决结果**：法院认定其不属于入户盗窃，因为被盗场所主要用于商业经营，且与生活区明确隔离。

  4. **案例四**：
     - **具体情形**：被告入公共办公用途的建筑物进行盗窃，但未达到与外界相对隔离和家庭生活双重特征的场所。
     - **案例说明**：被告人王某某盗窃进入的场所是被害人所在单位的职工宿舍，主要用于职工工作之余的休息，不具备家庭生活功能。
     - **判决结果**：法院认定其不属于入户盗窃，因为职工宿舍主要用于休息，不具备家庭生活属性。

  5. **案例五**：
     - **具体情形**：被害人不居住的长期搁置用房，被犯罪嫌疑人在这种场所中的盗窃行为。
     - **案例说明**：被告人朱某甲盗窃的场所是xx小区的一处房屋，该房屋自2012年起无人居住，不符合供家庭生活使用的特征。
     - **判决结果**：法院认定其不属于入户盗窃，因为该房屋未用于任何生活活动，不符合“户”的定义。

### 司法裁量

法官在具体案件中对“户”的定义的司法裁量主要基于以下几点：
- **实际用途**：确认被盗场所是否用于家庭生活。
- **场所性质**：确认场所有没有明确的与外界隔离的措施，如围墙、门等。
- **生活特征**：检查场所是否具备日常家庭生活的基本设施。
- **入侵方式**：查看被告人是否非法进入有明确隔离的家庭生活场所。
- **时间特征**：如家庭成员日常活动时间被侵入更容易被认定为“户”。

综上，“户”在独立案件中需结合案件场所的用途、隔离性和实际生活特征进行认定，司法审查需要严密把握其适用性以确保判决的法律公正。
\end{verbatim}

\subsubsection{English translation}

\begin{spverbatim}
Article 264 of the Criminal Law of the People's Republic of China states: "Theft of public or private property in large amounts" pertains to the conviction and sentencing of theft offenses. Implementing this article involves a potential ambiguity in the legal interpretation of the term "dwelling" in the context of "burglary into a dwelling." In judicial procedures, judges need to concretely define and interpret the concept of "dwelling" based on the specific facts of each case.

### Analysis

1. **Basic Definition**:
    - According to the "Interpretation on the Application of Law in Handling Theft Criminal Cases" by the Supreme People's Court and the Supreme People's Procuratorate, the concept of "dwelling" is characterized by two aspects: providing for family life and being relatively isolated from the outside world.
    - "Dwelling" typically includes residential premises, enclosed courtyards, rented homes used for living, etc.
    - Theft involving illegal entry into someone else's residence, which is isolated from the outside world, should be considered "burglary into a dwelling."

2. **Specific Application**:
    - When applying the concept of "dwelling," it is necessary to check whether the stolen premises are used for family life and are isolated from the outside.
    - Public places, commercial premises, or areas without clear isolation are generally not considered "dwellings."
    - In specific cases, the judge will assess factors such as the purpose of the property, the manner of intrusion, and the time of the offense.

3. **Exclusions**:
    - Not fitting the "living purpose": Such as shops used solely for business, public offices, etc.
    - Lack of "relative isolation": For example, places with no enclosure or highly open areas.
    - Consent or vacancy: If theft occurred with permission or in vacant, uninhabited buildings under renovation.

### Example Cases

- **Cases that meet the definition of "dwelling"**:

  1. **Case 1**:
     - **Facts**: The defendant illegally entered a private residence used for family living.
     - **Explanation**: The defendant, Mr. Yu, unlawfully entered Room 305 of Building XXX in Village XXX, District XX, which is rented by another person and used for family life, isolated from the outside world. This meets the definition of "dwelling."
     - **Verdict**: The court ruled it as burglary into a dwelling, as the defendant unlawfully entered a private residence that was relatively isolated.

  2. **Case 2**:
     - **Facts**: The defendant entered a family home repeatedly while the residents were absent.
     - **Explanation**: The defendant, Mr. Li, committed two thefts in the victim's residence, which was used for family life and isolated from the outside. This meets the definition of "dwelling."
     - **Verdict**: The court ruled it as burglary into a dwelling because the defendant illegally entered a residential property used for family living.

  3. **Case 3**:
     - **Facts**: The defendant climbed over a wall to enter a family courtyard isolated from the outside world and then committed theft.
     - **Explanation**: The defendant, Mr. Tian, illegally entered several victims' homes late at night. These homes were used for family life and were isolated from the outside world.
     - **Verdict**: The court ruled it was burglary into a dwelling because the defendant unlawfully entered an enclosed family living space.

  4. **Case 4**:
     - **Facts**: The defendant used tools to pry open a lock and break into a private residence to commit theft.
     - **Explanation**: The defendant, Mr. Zhang, used stolen keys to enter the victim's home to commit theft. This residence was used for family life and isolated from the outside.
     - **Verdict**: The court ruled it as burglary into a dwelling because the defendant unlawfully entered a private home.

  5. **Case 5**:
     - **Facts**: The defendant entered a residential area during a time when family members frequently came and went.
     - **Explanation**: The defendant, Mr. Wang, repeatedly stole property from a family residence using secretive methods. His actions occurred inside the victim's home, which was a residential space.
     - **Verdict**: The court ruled it as burglary into a dwelling because the defendant illegally entered a residential area used for family life and isolated from the outside.

- **Cases that do not meet the definition of "dwelling"**:

  1. **Case 1**:
     - **Facts**: The defendant stole property from a commercial store or in a public area.
     - **Explanation**: The defendant, Mr. Liu, committed theft in a shop operated by the victim, which was not a family residence.
     - **Verdict**: The court ruled it was not burglary into a dwelling because the shop was primarily for commercial use, not for family living.

  2. **Case 2**:
     - **Facts**: The defendant committed theft in an uninhabited property under renovation.
     - **Explanation**: The defendant, Mr. Zhang, stole from a public space used for vehicle parking in a building that was not a residential area.
     - **Verdict**: The court ruled it was not burglary into a dwelling because the property was not used for living purposes.

  3. **Case 3**:
     - **Facts**: The defendant committed theft in a commercial space that did not serve residential purposes.
     - **Explanation**: The defendant, Mr. Liu, entered a shop (labeled "China Mobile/ Yue Lu/ Designated Specialty Store") on the first floor of a building owned by the victim. The front area of the first floor was a commercial section selling mobile phones, while the rear and upper floors were residential areas. At the time of the offense, the commercial and residential areas were clearly separated by walls and doors.
     - **Verdict**: The court ruled it was not burglary into a dwelling because the stolen property was in a commercial space, separate from the residential area.

  4. **Case 4**:
     - **Facts**: The defendant entered a public office building to commit theft, but the location did not have the characteristics of a dwelling.
     - **Explanation**: The defendant, Mr. Wang, entered the dormitory of the victim's workplace, which employees used for rest, not for family living.
     - **Verdict**: The court ruled it was not burglary into a dwelling because the dormitory was used for rest and not for family living.

  5. **Case 5**:
     - **Facts**: The defendant stole from a long-term uninhabited property.
     - **Explanation**: The defendant, Mr. Zhu, committed theft in a house in the XX community that had been uninhabited since 2012 and was not used for family living.
     - **Verdict**: The court ruled it was not burglary into a dwelling because the property was not used for living activities and did not meet the definition of "dwelling."

### Judicial Discretion

Judges' judicial discretion in defining "dwelling" in specific cases mainly relies on the following factors:
- **Actual Use**: Confirming whether the stolen property was used for family life.
- **Nature of the Residence**: Confirming whether the residence had clear isolation measures such as walls or doors.
- **Living Features**: Checking whether the premises had basic facilities for daily family life.
- **Intrusion Method**: Determining whether the defendant illegally entered a clearly isolated family living space.
- **Time Features**: For instance, when family members' daily activities are disrupted, it is more likely to be recognized as a "dwelling."

In conclusion, the definition of "dwelling" in individual cases needs to be based on the use, isolation, and actual living characteristics of the premises. Judicial review requires careful attention to ensure the proper legal application and fairness of the verdict.
\end{spverbatim}

\clearpage

\section{Prompts}
\label{appendix:prompt}
\subsection{Original text in Chinese}
\subsubsection{Prompt for determining
whether court view provides a specific reason}

\noindent法律语言具有模糊性，而司法程序是对立法语言的一个明晰过程。在部分案件中，法官会根据案件事实对法律条文中的模糊概念进行具体化并在裁判文书中的“法庭观点”部分给出认定理由。我们考虑法条“\{\{article\}\}”中的模糊概念“\{\{concept\}\}”。我将给你一段法庭观点，请你判断法庭观点中，是否存在具体的句子解释“\{\{concept\}\}”适用或不适用于该案件的原因。先输出你的判断理由，然后严格按照以下格式输出你的最终判断。如果法庭观点中存在解释“\{\{concept\}\}”是否适用的句子，输出“[[是]]”；否则，输出“[[否]]”。

\vspace{1em}
\noindent[法庭观点]

\noindent\{\{court view\}\}

\vspace{1em}

\subsubsection{Prompt for classifying whether concept $c$
applies or not}

\noindent法律语言具有模糊性，而司法程序是对立法语言的一个明晰过程，法官会根据案件事实对法律条文中的模糊概念进行具体化并在裁判文书中的“法庭观点”部分给出认定理由。我们考虑法条“\{\{article\}\}”中的模糊概念“\{\{concept\}\}”。我将给你一段裁判文书中的法庭观点，请你判断法官认为模糊概念“\{\{concept\}\}”是否适用于案件中的情况。先给出你的判断理由，然后严格按照以下格式输出你的最终判断：如果“\{\{concept\}\}”适用于案件中的情况，输出“[[是]]”；否则，输出“[[否]]”。

\vspace{1em}
\noindent[法庭观点]

\noindent\{\{court view\}\}

\subsubsection{Prompt for extracting reason $r$ from court view}
\label{prompt_extract_reason}
\noindent法律语言具有模糊性，而司法程序是对立法语言的一个明晰过程。法官会根据案件事实对法律条文中的模糊词进行具体化并在裁判文书中的“法庭观点”部分进行分析。在法条“\{\{article\}\}”中，模糊概念是“\{\{concept\}\}”。请你阅读裁判文书中的法庭观点，提取出法官对模糊概念的认定理由。理由包括对案件事实经过的分析和最后的结论。比如，如果模糊概念是“户”，你需要提取出法官认为案件中的场所满足或不满足“户”的理由是什么。

\vspace{1em}
\noindent[法庭观点]

\noindent\{\{court view\}\}

\subsubsection{Prompt for generating concept interpretation}
\label{appendix:prompt_generate}
\begin{verbatim}
法律语言具有模糊性，而司法程序是对立法语言的一个明晰过程。法官会根据案件事实对法律条文中的模糊概念进行具体化并在裁判文书中分析模糊概念是否适用。请你阅读给出的JSON数据，对法条中的模糊概念进行解释。其中，"法条"是待分析的模糊概念所属的法条。"模糊概念"是你需要生成解释的法律概念。"参考文本"是从许多裁判文书中提取出的解释模糊概念的文本。
{
    "法条": {{article}},
    "模糊概念": {{concept}}
    "参考文本": {{reasons}}
}
以下是一个概念解释的样例，请以相同的格式规范输出。
{{Interpretation Example}}
\end{verbatim}

\subsubsection{Prompt for assigning consistency scores}
\label{cs_prompt}
\noindent请你参考法庭观点中对“\{\{crime\}\}”中的模糊概念“\{\{concept\}\}”的认定理由，对下面模型生成的认定理由的一致性进行1-10的打分。1分代表模型生成的认定理由和法庭观点中理由完全不一致，10分代表模型生成的认定理由和法庭观点中理由完全一致。请你先输出打分理由，然后以下列格式输出你的分数：[[n]]，其中n为你的分数。

\vspace{1em}

\noindent[模型生成的理由]

\noindent\{\{generated reason\}\}

\vspace{1em}

\noindent[法庭观点中理由]

\noindent\{\{gold reason\}\}

\subsubsection{Prompt for completing Legal Concept Entailment task}
\label{appendix:entailment_prompt}
法律语言具有模糊性，而司法程序是对立法语言的一个明晰过程。法官会根据案件事实对法律条文中的模糊概念进行具体化并在裁判文书中的“法庭观点”部分分析模糊概念是否适用。在法条“\{\{article\}\}”中，模糊概念是“\{\{concept\}\}”。请你阅读下面对模糊概念的解释，根据裁判文书中的事实描述，判断案件中的情况是否适用于模糊概念“\{\{concept\}\}”。先提供判定理由，然后严格按照以下格式输出你的最终判断：如果符合模糊概念“\{\{concept\}\}”的定义，输出“[[是]]”，否则输出“[[否]]”。

\vspace{1em}

\noindent[模糊概念的解释]

\noindent\{\{interpretation\}\}

\vspace{1em}

\noindent[事实描述]

\noindent\{\{fact\}\}

\subsection{English translation}
\subsubsection{Prompt for determining whether court view provides a specific reason}

\noindent Legal language is inherently vague, and the judicial process serves as a clarification of legislative language. In some cases, judges may concretize vague terms in the legal texts based on the facts of the case and provide reasons for their determination in the "court view" section of the ruling document. We consider the vague concept "\{\{concept\}\}" in the legal article "\{\{article\}\}". I will give you a segment of the court view; please determine whether there is a specific sentence in the court view that explains the reason why "\{\{concept\}\}" does or does not apply to the case. First, output your reasoning for the judgment, then strictly follow the format below for your final conclusion. If there is a sentence explaining whether "\{\{concept\}\}" applies, output "[[Yes]]"; otherwise, output "[[No]]".

\vspace{1em}
\noindent[Court View]

\noindent\{\{court view\}\}

\vspace{1em}

\subsubsection{Prompt for classifying whether concept $c$ applies or not}

\noindent Legal language is inherently vague, and the judicial process serves as a clarification of legislative language, where judges can concretize vague terms in legal texts based on the facts of the case and provide reasons for their determination in the "court view" section of the ruling document. We consider the vague concept "\{\{concept\}\}" in the legal article "\{\{article\}\}". I will give you a segment of the court view; please determine whether the judge believes the vague concept "\{\{concept\}\}" applies to the situation in the case. First, provide your reasoning for the judgment, then strictly follow the format below for your final conclusion: If "\{\{concept\}\}" applies to the situation in the case, output "[[Yes]]"; otherwise, output "[[No]]".

\vspace{1em}
\noindent[Court View]

\noindent\{\{court view\}\}

\subsubsection{Prompt for extracting reason $r$ from court view}
\label{prompt_extract_reason_en}
\noindent Legal language is inherently vague, and the judicial process serves as a clarification of legislative language. Judges can concretize vague terms in legal texts based on the facts of the case and analyze them in the "court view" section of the ruling document. In the legal article "\{\{article\}\}", the vague concept is "\{\{concept\}\}". Please read the court view in the ruling document and extract the judge's reasoning for the determination of the vague concept. The reasoning includes the analysis of the facts of the case and the final conclusion. For example, if the vague concept is "dwelling," you need to extract the reasons why the judge believes the place in the case satisfies or does not satisfy the "dwelling" criterion.

\vspace{1em}
\noindent[Court View]

\noindent\{\{court view\}\}

\subsubsection{Prompt for generating concept interpretation}
\label{appendix:prompt_generate_en}
\begin{spverbatim}
Legal language is inherently vague, and the judicial process serves as a clarification of legislative language. Judges can concretize vague terms in legal texts based on the facts of the case and analyze whether the vague concept applies in the ruling document. Please read the given JSON data and interpret the vague concept in the legal article. Among them, "article" is the legal article to which the vague concept belongs. "vague concept" is the legal concept you need to interpret. "Reference text" is the text extracted from many ruling documents explaining the vague concept.
{
    "Article": {{article}},
    "vague concept": {{concept}}
    "Reference text": {{reasons}}
}
Below is an example of a concept interpretation. Please format your output following the same standard.
{{Interpretation Example}}
\end{spverbatim}

\subsubsection{Prompt for assigning consistency scores}
\label{cs_prompt_en}
\noindent Please refer to the reasons for determining the vague concept "\{\{concept\}\}" in "\{\{crime\}\}" from the court view and rate the consistency of the following model-generated reasons on a scale of 1-10. A score of 1 indicates that the model-generated reasons are completely inconsistent with the reasons in the court view, while a score of 10 indicates complete consistency. First, output your reasoning for the score, then output your score in the following format: [[n]], where n is your score.

\vspace{1em}

\noindent[Model-generated Reason]

\noindent\{\{generated reason\}\}

\vspace{1em}

\noindent[Reason in Court View]

\noindent\{\{gold reason\}\}

\subsubsection{Prompt for completing Legal Concept Entailment task}
\label{appendix:entailment_prompt_en}
Legal language is inherently vague, and the judicial process serves as a clarification of legislative language. Judges can concretize vague terms in legal texts based on the facts of the case and analyze them in the "court view" section of the ruling document to determine whether the vague concept applies. In the legal article "\{\{article\}\}", the vague concept is "\{\{concept\}\}". Please read the following interpretation of the vague concept, and based on the factual description in the ruling document, determine whether the situation in the case applies to the vague concept "\{\{concept\}\}". First, provide reasons for your determination, then strictly follow the format below for your final conclusion: If it meets the definition of the vague concept "\{\{concept\}\}", output "[[Yes]]"; otherwise, output "[[No]]".

\vspace{1em}

\noindent[Interpretation of vague Concept]

\noindent\{\{interpretation\}\}

\vspace{1em}

\noindent[Factual Description]

\noindent\{\{fact\}\}

\section{Details of vague concepts}
\label{appendix:concepts}
Table \ref{table:dataset} presents the detailed statistics of the test dataset for the legal concept entailment task.
Tables \ref{tab:concepts1} and \ref{tab:concepts2} present the vague concepts we interpret and their corresponding legal articles.

\begin{table*}[h]
    \centering
    \scalebox{1}{
    \begin{tabular}{lr}
        \toprule
        Test Dataset &  \\ 
        \midrule
        \# Concepts & 16  \\
        \# Cases & 2652 \\
        \quad - positive & 1714 \\
        \quad - negative & 837 \\
        
        \# Average court view length & 653.1 \\
        \# Average fact length & 4787.9 \\
        \# Average reason length & 160.5 \\
        
        \bottomrule
    \end{tabular}
    }
    \caption{Basic statistics of the test dataset.}
    \label{table:dataset}
\end{table*}

\begin{table*}[h]
  \centering
   \scalebox{0.68}{

   \begin{tabular}{p{3cm}p{18cm}}
    \toprule
    \textbf{Vague concept} & \textbf{Article} \\ \midrule
    情节严重 & 第一百二十五条：非法制造、买卖、运输、邮寄、储存枪支、弹药、爆炸物的，处三年以上十年以下有期徒刑；情节严重的，处十年以上有期徒刑、无期徒刑或者死刑。非法制造、买卖、运输、储存毒害性、放射性、传染病病原体等物质，危害公共安全的，依照前款的规定处罚。单位犯前两款罪的，对单位判处罚金，并对其直接负责的主管人员和其他直接责任人员，依照第一款的规定处罚。 \\ \midrule
情节严重 & 第一百二十八条：违反枪支管理规定，非法持有、私藏枪支、弹药的，处三年以下有期徒刑、拘役或者管制；情节严重的，处三年以上七年以下有期徒刑。依法配备公务用枪的人员，非法出租、出借枪支的，依照前款的规定处罚。依法配置枪支的人员，非法出租、出借枪支，造成严重后果的，依照第一款的规定处罚。单位犯第二款、第三款罪的，对单位判处罚金，并对其直接负责的主管人员和其他直接责任人员，依照第一款的规定处罚。 \\ \midrule
逃逸 & 第一百三十三条：违反交通运输管理法规，因而发生重大事故，致人重伤、死亡或者使公私财产遭受重大损失的，处三年以下有期徒刑或者拘役；交通运输肇事后逃逸或者有其他特别恶劣情节的，处三年以上七年以下有期徒刑；因逃逸致人死亡的，处七年以上有期徒刑。在道路上驾驶机动车，有下列情形之一的，处拘役，并处罚金：（一）追逐竞驶，情节恶劣的；（二）醉酒驾驶机动车的；（三）从事校车业务或者旅客运输，严重超过额定乘员载客，或者严重超过规定时速行驶的；（四）违反危险化学品安全管理规定运输危险化学品，危及公共安全的。机动车所有人、管理人对前款第三项、第四项行为负有直接责任的，依照前款的规定处罚。有前两款行为，同时构成其他犯罪的，依照处罚较重的规定定罪处罚。第一百三十三条之二 对行驶中的公共交通工具的驾驶人员使用暴力或者抢控驾驶操纵装置，干扰公共交通工具正常行驶，危及公共安全的，处一年以下有期徒刑、拘役或者管制，并处或者单处罚金。前款规定的驾驶人员在行驶的公共交通工具上擅离职守，与他人互殴或者殴打他人，危及公共安全的，依照前款的规定处罚。有前两款行为，同时构成其他犯罪的，依照处罚较重的规定定罪处罚。 \\ \midrule
严重情节 & 第二百二十四条：有下列情形之一，以非法占有为目的，在签订、履行合同过程中，骗取对方当事人财物，数额较大的，处三年以下有期徒刑或者拘役，并处或者单处罚金；数额巨大或者有其他严重情节的，处三年以上十年以下有期徒刑，并处罚金；数额特别巨大或者有其他特别严重情节的，处十年以上有期徒刑或者无期徒刑，并处罚金或者没收财产：（一）以虚构的单位或者冒用他人名义签订合同的；（二）以伪造、变造、作废的票据或者其他虚假的产权证明作担保的；（三）没有实际履行能力，以先履行小额合同或者部分履行合同的方法，诱骗对方当事人继续签订和履行合同的；（四）收受对方当事人给付的货物、货款、预付款或者担保财产后逃匿的；（五）以其他方法骗取对方当事人财物的。组织、领导以推销商品、提供服务等经营活动为名，要求参加者以缴纳费用或者购买商品、服务等方式获得加入资格，并按照一定顺序组成层级，直接或者间接以发展人员的数量作为计酬或者返利依据，引诱、胁迫参加者继续发展他人参加，骗取财物，扰乱经济社会秩序的传销活动的，处五年以下有期徒刑或者拘役，并处罚金；情节严重的，处五年以上有期徒刑，并处罚金。 \\ \midrule
合同 & 第二百二十四条：有下列情形之一，以非法占有为目的，在签订、履行合同过程中，骗取对方当事人财物，数额较大的，处三年以下有期徒刑或者拘役，并处或者单处罚金；数额巨大或者有其他严重情节的，处三年以上十年以下有期徒刑，并处罚金；数额特别巨大或者有其他特别严重情节的，处十年以上有期徒刑或者无期徒刑，并处罚金或者没收财产：（一）以虚构的单位或者冒用他人名义签订合同的；（二）以伪造、变造、作废的票据或者其他虚假的产权证明作担保的；（三）没有实际履行能力，以先履行小额合同或者部分履行合同的方法，诱骗对方当事人继续签订和履行合同的；（四）收受对方当事人给付的货物、货款、预付款或者担保财产后逃匿的；（五）以其他方法骗取对方当事人财物的。组织、领导以推销商品、提供服务等经营活动为名，要求参加者以缴纳费用或者购买商品、服务等方式获得加入资格，并按照一定顺序组成层级，直接或者间接以发展人员的数量作为计酬或者返利依据，引诱、胁迫参加者继续发展他人参加，骗取财物，扰乱经济社会秩序的传销活动的，处五年以下有期徒刑或者拘役，并处罚金；情节严重的，处五年以上有期徒刑，并处罚金。 \\ \midrule
非法占有为目的 & 第二百二十四条：有下列情形之一，以非法占有为目的，在签订、履行合同过程中，骗取对方当事人财物，数额较大的，处三年以下有期徒刑或者拘役，并处或者单处罚金；数额巨大或者有其他严重情节的，处三年以上十年以下有期徒刑，并处罚金；数额特别巨大或者有其他特别严重情节的，处十年以上有期徒刑或者无期徒刑，并处罚金或者没收财产：（一）以虚构的单位或者冒用他人名义签订合同的；（二）以伪造、变造、作废的票据或者其他虚假的产权证明作担保的；（三）没有实际履行能力，以先履行小额合同或者部分履行合同的方法，诱骗对方当事人继续签订和履行合同的；（四）收受对方当事人给付的货物、货款、预付款或者担保财产后逃匿的；（五）以其他方法骗取对方当事人财物的。组织、领导以推销商品、提供服务等经营活动为名，要求参加者以缴纳费用或者购买商品、服务等方式获得加入资格，并按照一定顺序组成层级，直接或者间接以发展人员的数量作为计酬或者返利依据，引诱、胁迫参加者继续发展他人参加，骗取财物，扰乱经济社会秩序的传销活动的，处五年以下有期徒刑或者拘役，并处罚金；情节严重的，处五年以上有期徒刑，并处罚金。 \\ 
    \bottomrule
    \end{tabular}
}
  \caption{The 16 vague concepts and their corresponding articles used in our study. (i)}

  \label{tab:concepts1}
\end{table*}

\begin{table*}[p]
  \centering
   \scalebox{0.7}{

   \begin{tabular}{p{3cm}p{16cm}}
    \toprule
    \textbf{Vague concept} & \textbf{Article} \\ \midrule
    
情节严重 & 第二百二十五条：违反国家规定，有下列非法经营行为之一，扰乱市场秩序，情节严重的，处五年以下有期徒刑或者拘役，并处或者单处违法所得一倍以上五倍以下罚金；情节特别严重的，处五年以上有期徒刑，并处违法所得一倍以上五倍以下罚金或者没收财产：（一）未经许可经营法律、行政法规规定的专营、专卖物品或者其他限制买卖的物品的；（二）买卖进出口许可证、进出口原产地证明以及其他法律、行政法规规定的经营许可证或者批准文件的；（三）未经国家有关主管部门批准非法经营证券、期货、保险业务的，或者非法从事资金支付结算业务的；（四）其他严重扰乱市场秩序的非法经营行为。 \\ \midrule
户 & 第二百六十四条：盗窃公私财物，数额较大的，或者多次盗窃、入户盗窃、携带凶器盗窃、扒窃的，处三年以下有期徒刑、拘役或者管制，并处或者单处罚金；数额巨大或者有其他严重情节的，处三年以上十年以下有期徒刑，并处罚金；数额特别巨大或者有其他特别严重情节的，处十年以上有期徒刑或者无期徒刑，并处罚金或者没收财产。 \\ \midrule
    职务 & 第二百七十一条：公司、企业或者其他单位的工作人员，利用职务上的便利，将本单位财物非法占为己有，数额较大的，处三年以下有期徒刑或者拘役，并处罚金；数额巨大的，处三年以上十年以下有期徒刑，并处罚金；数额特别巨大的，处十年以上有期徒刑或者无期徒刑，并处罚金。国有公司、企业或者其他国有单位中从事公务的人员和国有公司、企业或者其他国有单位委派到非国有公司、企业以及其他单位从事公务的人员有前款行为的，依照本法第三百八十二条、第三百八十三条的规定定罪处罚。 \\ \midrule
单位 & 第二百七十二条：公司、企业或者其他单位的工作人员，利用职务上的便利，挪用本单位资金归个人使用或者借贷给他人，数额较大、超过三个月未还的，或者虽未超过三个月，但数额较大、进行营利活动的，或者进行非法活动的，处三年以下有期徒刑或者拘役；挪用本单位资金数额巨大的，处三年以上七年以下有期徒刑；数额特别巨大的，处七年以上有期徒刑。国有公司、企业或者其他国有单位中从事公务的人员和国有公司、企业或者其他国有单位委派到非国有公司、企业以及其他单位从事公务的人员有前款行为的，依照本法第三百八十四条的规定定罪处罚。有第一款行为，在提起公诉前将挪用的资金退还的，可以从轻或者减轻处罚。其中，犯罪较轻的，可以减轻或者免除处罚。 \\ \midrule
情节严重 & 第二百八十条：伪造、变造、买卖或者盗窃、抢夺、毁灭国家机关的公文、证件、印章的，处三年以下有期徒刑、拘役、管制或者剥夺政治权利，并处罚金；情节严重的，处三年以上十年以下有期徒刑，并处罚金。伪造公司、企业、事业单位、人民团体的印章的，处三年以下有期徒刑、拘役、管制或者剥夺政治权利，并处罚金。伪造、变造、买卖居民身份证、护照、社会保障卡、驾驶证等依法可以用于证明身份的证件的，处三年以下有期徒刑、拘役、管制或者剥夺政治权利，并处罚金；情节严重的，处三年以上七年以下有期徒刑，并处罚金。在依照国家规定应当提供身份证明的活动中，使用伪造、变造的或者盗用他人的居民身份证、护照、社会保障卡、驾驶证等依法可以用于证明身份的证件，情节严重的，处拘役或者管制，并处或者单处罚金。有前款行为，同时构成其他犯罪的，依照处罚较重的规定定罪处罚。第二百八十条之二 盗用、冒用他人身份，顶替他人取得的高等学历教育入学资格、公务员录用资格、就业安置待遇的，处三年以下有期徒刑、拘役或者管制，并处罚金。组织、指使他人实施前款行为的，依照前款的规定从重处罚。国家工作人员有前两款行为，又构成其他犯罪的，依照数罪并罚的规定处罚。 \\ \midrule
情节严重 & 第三百一十二条：明知是犯罪所得及其产生的收益而予以窝藏、转移、收购、代为销售或者以其他方法掩饰、隐瞒的，处三年以下有期徒刑、拘役或者管制，并处或者单处罚金；情节严重的，处三年以上七年以下有期徒刑，并处罚金。单位犯前款罪的，对单位判处罚金，并对其直接负责的主管人员和其他直接责任人员，依照前款的规定处罚。 \\ \midrule
情节严重 & 第三百四十八条：非法持有鸦片一千克以上、海洛因或者甲基苯丙胺五十克以上或者其他毒品数量大的，处七年以上有期徒刑或者无期徒刑，并处罚金；非法持有鸦片二百克以上不满一千克、海洛因或者甲基苯丙胺十克以上不满五十克或者其他毒品数量较大的，处三年以下有期徒刑、拘役或者管制，并处罚金；情节严重的，处三年以上七年以下有期徒刑，并处罚金。 \\ \midrule
情节严重 & 第三百五十九条：引诱、容留、介绍他人卖淫的，处五年以下有期徒刑、拘役或者管制，并处罚金；情节严重的，处五年以上有期徒刑，并处罚金。引诱不满十四周岁的幼女卖淫的，处五年以上有期徒刑，并处罚金。 \\ \midrule
情节严重 & 第三百八十四条：国家工作人员利用职务上的便利，挪用公款归个人使用，进行非法活动的，或者挪用公款数额较大、进行营利活动的，或者挪用公款数额较大、超过三个月未还的，是挪用公款罪，处五年以下有期徒刑或者拘役；情节严重的，处五年以上有期徒刑。挪用公款数额巨大不退还的，处十年以上有期徒刑或者无期徒刑。挪用用于救灾、抢险、防汛、优抚、扶贫、移民、救济款物归个人使用的，从重处罚。 \\ \midrule
情节严重 & 第三百九十条：对犯行贿罪的，处五年以下有期徒刑或者拘役，并处罚金；因行贿谋取不正当利益，情节严重的，或者使国家利益遭受重大损失的，处五年以上十年以下有期徒刑，并处罚金；情节特别严重的，或者使国家利益遭受特别重大损失的，处十年以上有期徒刑或者无期徒刑，并处罚金或者没收财产。行贿人在被追诉前主动交待行贿行为的，可以从轻或者减轻处罚。其中，犯罪较轻的，对侦破重大案件起关键作用的，或者有重大立功表现的，可以减轻或者免除处罚。为谋取不正当利益，向国家工作人员的近亲属或者其他与该国家工作人员关系密切的人，或者向离职的国家工作人员或者其近亲属以及其他与其关系密切的人行贿的，处三年以下有期徒刑或者拘役，并处罚金；情节严重的，或者使国家利益遭受重大损失的，处三年以上七年以下有期徒刑，并处罚金；情节特别严重的，或者使国家利益遭受特别重大损失的，处七年以上十年以下有期徒刑，并处罚金。单位犯前款罪的，对单位判处罚金，并对其直接负责的主管人员和其他直接责任人员，处三年以下有期徒刑或者拘役，并处罚金。 \\ 
    \bottomrule
    \end{tabular}
}
  \caption{The 16 vague concepts and their corresponding articles used in our study. (ii)}

  \label{tab:concepts2}
\end{table*}

\end{CJK*}
\end{document}